\documentclass[a4paper,fleqn]{cas-dc}

\usepackage[numbers]{natbib}
\usepackage{subcaption}
\usepackage{caption}

\def\tsc#1{\csdef{#1}{\textsc{\lowercase{#1}}\xspace}}
\tsc{WGM}
\tsc{QE}

\begin{document}
\let\WriteBookmarks\relax
\def\floatpagepagefraction{1}
\def\textpagefraction{.001}

\shorttitle{DiDrive}

\shortauthors{Q. Guo et~al.}  

\title [mode = title]{DiDrive: A Risk-Aware Hierarchical Diffusion Framework for Safe Offline Reinforcement Learning in Autonomous Driving}  


\tnotetext[1]{This work was supported by the Natural Science Foundation of Fujian Province, China under Grant 2026J001253.}

%

\author[1]{Qisong Guo}

\author[2]{Jingtang Chen}

\author[1]{Zhilin Chen}

\author[3]{Pei Xu}

\author[1]{Mingjian Fu}
\cormark[1]
\ead{sinceway@fzu.edu.cn}

\author[1]{Wenxi Liu}

\author[1]{Yuanlong Yu}





\affiliation[1]{organization={College of Computer and Data Science, Fuzhou University},
            addressline={}, 
            city={Fuzhou},
            postcode={350116}, 
            state={Fujian},
            country={China}}






\affiliation[2]{organization={Maynooth International College of Engineering, Fuzhou University},
            addressline={}, 
            city={Fuzhou},
            postcode={350116}, 
            state={Fujian},
            country={China}}

\affiliation[3]{organization={Institute of Automation, Chinese Academy of Sciences},
            addressline={}, 
            city={Beijing},
            postcode={100190}, 
            state={},
            country={China}}
\cortext[1]{Corresponding author}



\begin{abstract}
Autonomous driving requires safe and reliable sequential decision-making in highly dynamic traffic environments. Offline reinforcement learning (RL) has emerged as a promising paradigm by learning directly from static datasets, thereby mitigating online exploration hazards. However, applying offline RL faces significant bottlenecks. While diffusion models effectively capture multimodal behavioral priors, they are susceptible to distribution shift and heavy-tailed risk signals, which can distort the reverse denoising direction and generate out-of-distribution (OOD) or hazardous actions. This vulnerability is further compounded in complex traffic scenarios, where high-dimensional state inputs often burden agents with redundant environmental information, masking essential local risks and misguiding action generation.

To address these intertwined challenges, we propose DiDrive, a distribution-guided offline diffusion framework for safe RL. DiDrive is underpinned by two closely integrated core components: the Risk-Aware Hierarchical Diffusion (RHDif) architecture and the Distribution Correction Estimation with Diffusion for Driving (3DICE) policy optimization paradigm. In the state space, RHDif manages input complexity; it employs a low-level risk-gated encoder to enhance local danger features and a high-level contextual modulator to align global semantics. The fused features from these modules are injected into successive reverse denoising steps as a joint condition, guiding the model to filter redundancy and focus on core threats. In the action space, 3DICE builds a four-stage mechanism with internalized safety. Building upon the refined representations extracted by RHDif, it further mitigates OOD overestimation and gradient oscillation through in-sample calibrated guidance via density-ratio weighting, spatiotemporal collaborative optimization, and ensemble-based candidate ranking. Working in tandem, RHDif refines environmental perception while 3DICE constrains action uncertainty, enhancing the overall safety utility of the hierarchical design.

Experiments on the CARLA simulation benchmark demonstrate DiDrive's stability across varying traffic densities. Compared to baselines including IQL, CQL, and Diffusion-QL, DiDrive exhibits improved overall performance in success rate, average reward, and lane deviation. Notably, in complex high-density scenarios with 60 background vehicles, DiDrive achieved the highest success rate of 85\% and the highest average reward of 4295.68. By effectively integrating risk-aware representation with distribution correction optimization, this study provides a robust and promising pathway for safe autonomous driving decision-making in highly complex environments.

\end{abstract}




\maketitle

\section{Introduction}
\label{sec:intro}

Safe and accurate sequential decision-making in high-dimensional, dynamic, and uncertain traffic environments is the bedrock of autonomous driving \cite{Wen_2024_ICLR_DiLu,Hu_2023_CVPR_UniAD,Zheng_2024_ECCV_GenAD}. While data-driven approaches like imitation learning and online reinforcement learning have driven significant progress \cite{10602544, 9904958}, they face critical bottlenecks in building highly reliable closed-loop systems. Imitation learning often fails to cover long-tailed risks, whereas online reinforcement learning introduces unacceptable safety hazards during trial-and-error exploration. As a result, offline reinforcement learning has emerged as a vital alternative, learning effective policies directly from static datasets to circumvent the risks and costs of online interaction \cite{levine2020offline}.

Despite its promise, offline reinforcement learning is fundamentally bottlenecked by distribution shift \cite{10078377}. Current solutions generally fall into three categories, each with distinct limitations in the context of autonomous driving. The first technical route relies on policy constraints and value regularization \cite{Ran2023PolicyRW,Kostrikov2021OfflineRL,kumar2019bear,NEURIPS2020_0d2b2061}. Although these methods reduce distributional mismatch, their unimodal Gaussian assumptions struggle to capture the complex, multimodal nature of driving behaviors, such as yielding and overtaking. The second technical route leverages highly expressive generative models, notably diffusion models \cite{NEURIPS2023_3b3889d3,NEURIPS2023_d45e0bfb,zheng2025diffusionbased,11094573,11093059,chi2025diffusionpolicy}. However, because the state-action space in autonomous driving is high dimensional and continuous, diffusion-based policies are prone to sampling actions outside the training data support. Relying on value functions to filter these out-of-distribution candidate actions often leads to overestimation, potentially misguiding the agent into catastrophic decisions. The third technical route employs distribution correction estimation, bypassing explicit out-of-distribution queries through in-sample importance weighting. Yet, in driving scenarios dominated by extreme, heavy-tailed risk signals like collision penalties \cite{pmlr-v238-zhu24a}, traditional correction methods suffer from severe numerical instability, causing gradient oscillation and distorted policy guidance.

Beyond challenges in the action space, high-dimensional state representation poses another significant hurdle. Complex traffic environments consist of road backgrounds, static obstacles, dynamic participants, and potential conflict zones \cite{Jiang2023VADVS}. Existing methods typically process these elements without adequate differentiation. This uniform treatment diminishes the model's ability to efficiently leverage safety-critical information and delays the identification of high-risk targets during sudden, hazardous events \cite{Jia_2023_ICCV_DriveAdapter, Jia_2023_CVPR_ThinkTwice}.

To overcome these intertwined challenges, policy optimization for autonomous driving must first restrict action queries to supported data regions. Furthermore, it requires strict risk prioritization when processing complex environmental inputs. To this end, models should maximally exploit the state-action samples already present in the static dataset. By assigning greater weight to high-quality decision samples during training, the diffusion denoising process is naturally steered toward a superior action distribution within the data support. Simultaneously, the agent must filter out high-dimensional background redundancy to accurately isolate critical safety cues. To seamlessly integrate support-aware policy optimization with risk-aware state representation, we propose DiDrive, a distribution-guided diffusion framework for safe offline reinforcement learning.

At the core of DiDrive is the Distribution Correction Estimation with Diffusion for Driving (3DICE) paradigm, which structures policy optimization into four stages: representation, guidance, optimization, and selection. Under this paradigm, rather than explicitly querying unseen actions, the diffusion model adjusts its reverse denoising trajectory using in-sample guidance derived from historical experience weights. Coupled with progressive parameter integration (PPI), 3DICE securely shifts the behavior policy toward high-quality, in-support actions, effectively mitigating value overestimation and the instability caused by heavy-tailed rewards.

To combat information overload and the lack of risk differentiation in the state space, we design the Risk-Aware Hierarchical Diffusion (RHDif) architecture. At the lower level, a Risk-Gated Spatiotemporal Encoder (RGSE) actively amplifies the feature responses of critical threats via bottom-up local risk perception when safety margins shrink. Concurrently, the high-level module utilizes a Cross-Modal Contextual Modulator (CMCM) to execute top-down contextual filtering, effectively suppressing irrelevant background interference.

The main contributions of this work are threefold:
\begin{itemize}
\item First, we propose the 3DICE paradigm, formulating a novel optimization framework that integrates diffusion priors, in-sample guidance, spatiotemporal collaborative optimization, and internalized support-set constraints. This design effectively avoids value overestimation from out-of-distribution queries and stabilizes training against heavy-tailed risk signals.
\item Second, we design the RHDif architecture to decouple local risk perception from high-dimensional semantic denoising. By synergistically combining the RGSE with the CMCM, the model efficiently filters redundant noise to focus computational resources on core threats.
\item Third, extensive evaluations in the high-fidelity CARLA simulator demonstrate that DiDrive significantly outperforms state-of-the-art baselines across long-tailed, high-density, and safety-critical scenarios, proving its robust generalization and decision reliability.
\end{itemize}

\section{Related Work}

\subsection{Offline Reinforcement Learning}

Offline reinforcement learning aims to learn decision-making policies directly from static datasets without additional environment interaction. This paradigm is particularly valuable in applications where data collection is costly or safety requirements are stringent. Existing studies related to this work can be broadly categorized into three progressive directions: conservative value and policy learning, expressive policy modeling, and distribution correction.

Representative conservative learning methods include behavior-constrained algorithms like BCQ \cite{Fujimoto2018OffPolicyDR} and TD3-BC \cite{Fujimoto2021TD3BC}, as well as value regularization methods like CQL \cite{NEURIPS2020_0d2b2061} and implicit learning via IQL \cite{Kostrikov2021OfflineRL}. BCQ mitigates extrapolation error by constraining the learned policy to select actions close to the behavior policy, while CQL imposes conservative constraints on value estimates of OOD actions to reduce overestimation. The empirical study by the Waymo team \cite{GuillenPerez2024OfflineRL} further validates the effectiveness of CQL in autonomous driving, showing that it can leverage suboptimal data to improve policy performance and outperform traditional behavior cloning on medium-scale datasets.

While these existing offline RL methods effectively mitigate OOD overestimation, they typically rely on relatively restrictive policy parameterizations, such as unimodal Gaussian distributions. This substantially limits their ability to model the highly multimodal nature of human driving behaviors. To overcome this limitation, more expressive policy representations have been explored. For instance, DT \cite{NEURIPS2021_7f489f64} formulates offline RL as return-conditioned sequence modeling with a Transformer architecture, while Diffusion-QL \cite{Wang2022DiffusionPA} and QGPO \cite{10.5555/3618408.3619356} employ diffusion models to enhance the multimodal expressiveness of policy distributions.

Another line of work focuses on distribution correction in offline reinforcement learning. Its theoretical foundation can be traced back to the DICE family of methods, which estimate stationary distribution ratios for off-policy evaluation and optimization. DualDICE estimates the discounted stationary occupancy ratio between the target policy and the behavior distribution \cite{nachum2019dualdice}, while GenDICE extends this idea to general stationary distribution correction through variational divergence minimization \cite{zhang2020gendice}. Compared with methods that directly search for high-value actions in continuous action spaces, distribution correction methods reduce reliance on explicit value estimates of unseen actions, making them theoretically suitable for alleviating OOD queries.

Although these diverse offline RL methods have shown promise in standard benchmarks, their direct application to autonomous driving reveals critical gaps in policy optimization. Specifically, standard value regularization and distribution correction techniques are highly sensitive to the heavy-tailed reward distributions typical of driving datasets, such as severe collision penalties. When processing these heavy-tailed signals, they often suffer from pronounced numerical instability and gradient oscillation, which distorts policy guidance. Consequently, there is a pressing need for a robust optimization mechanism that can stabilize in-sample value estimation against extreme safety risks, thereby laying a reliable foundation for integrating more expressive generative policies.

\subsection{Application of Diffusion Models in Autonomous Driving}

As an emerging generative modeling approach, diffusion models can effectively capture multimodal driving behavior distributions through progressive denoising \cite{Ho2020DDPM,Song2021ScoreSDE,Dhariwal2021GuidedDiffusion,zheng2025diffusionbased,Zheng_2024_ECCV_GenAD}, providing a promising technical pathway for handling uncertainty in autonomous driving. Recent studies have demonstrated their potential in autonomous driving behavior planning \cite{11094573,11093059}. Compared with traditional generative models, diffusion models are more expressive in representing complex probability distributions and can generate diverse, reasonable, and smooth driving trajectories or behavioral sequences, thereby addressing multimodality and uncertainty in complex traffic scenarios.

In behavior planning, diffusion models have achieved notable progress. The pioneering work PlanningWD by Janner et al. \cite{Janner2022PlanningWD} validated the effectiveness of diffusion models in sequential decision-making, and subsequent studies have extended this idea to prediction, simulation, and end-to-end planning for autonomous driving \cite{Hu_2023_CVPR_UniAD}. At the multi-agent interaction level, MotionDiffuser by Jiang et al. \cite{10203706} replaces traditional regression-based prediction with diffusion-based trajectory distribution modeling, achieving controllable generation with interaction consistency and multimodal diversity through a permutation-invariant Transformer architecture. Diffusion-ES further combines trajectory denoising with gradient-free evolutionary search for closed-loop autonomous-driving planning and instruction-guided trajectory optimization \cite{Yang2024DiffusionES}. For real-time end-to-end driving, DiffusionDrive by Liao et al. \cite{11094573} introduces a Truncated Diffusion Policy, which uses predefined multimodal anchors as priors and reduces the denoising process to only two steps, achieving 45 FPS inference while preserving trajectory diversity. In city-level simulation, SceneDiffuser++ by Jeon et al. \cite{Tan2025SceneDiffuserCT} proposes CitySim based on a generative world model to jointly model dynamic agent generation, removal, and traffic light logic. DriveDreamer-2 further explores LLM-enhanced world modeling for diverse driving-video generation, showing the potential of generative world models for synthesizing uncommon and safety-critical driving scenarios \cite{Zhao2025DriveDreamer2}. To further reduce planning latency, DiffuserLite by Dong et al. \cite{NEURIPS2024_dd6a47bc} adopts a coarse-to-fine Progressive Refinement Strategy (PRP) to filter distant redundant information, increasing the decision frequency to 122 Hz and demonstrating the deployment potential of diffusion models on embedded vehicle platforms.

However, existing diffusion architectures still face limitations in autonomous driving. First, they may struggle to fully capture the multimodal state characteristics of complex traffic environments. Second, the diffusion sampling process can generate actions or trajectory segments outside the support of the training data distribution \cite{NEURIPS2024_a9f3457f}. If value functions are then used to filter these candidates, out-of-distribution value estimation errors may be further amplified.

Furthermore, in terms of risk perception and representation learning, complex traffic scenes contain substantial background information that is irrelevant to the current decision. Existing standard diffusion architectures often process high-dimensional environmental inputs uniformly, lacking fine-grained spatiotemporal attention mechanisms to dynamically quantify risk indicators. Consequently, these methods fail to effectively address the inherent contradiction between amplifying local safety-critical risk signals, such as vehicle cut-ins or emergency braking, and suppressing global pseudo-risks, such as distant static obstacles or non-interacting vehicles in opposite lanes \cite{RACP_2024, DQ_GAT_2022}. This deficiency in risk-differentiated representation causes the modeling capacity to be dispersed across irrelevant inputs, frequently leading to overly conservative driving behaviors or erratic decision-making. Therefore, for safety-critical events, the model must enhance its responsiveness to key local risk factors while relying on global semantic consistency to filter out pseudo-risk interference.

\section{Methodology}

\par\vspace{1em}
\begin{center}
\begin{minipage}{0.95\columnwidth}
    \centering
    \includegraphics[width=\linewidth]{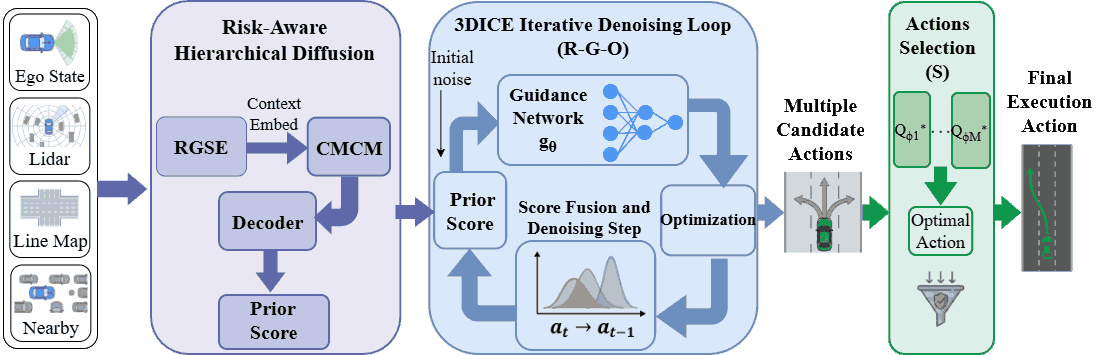}
    \captionof{figure}{Overall architecture of DiDrive.}
    \label{fig:didrive}
\end{minipage}
\end{center}

This section presents the proposed DiDrive (Diffusion-based Driver) framework, as illustrated in Fig.~\ref{fig:didrive}. Designed for safety-critical domains such as autonomous driving, DiDrive tackles the challenges posed by high-dimensional state spaces, complex temporal dependencies, and highly imbalanced long-tailed data distributions. Under such conditions, traditional offline RL methods often struggle to capture multimodal driving behaviors and are susceptible to value overestimation on OOD actions. To overcome these bottlenecks, DiDrive synergistically couples the RHDif architecture with stabilized distribution correction mechanisms, thereby refining the offline policy optimization paradigm.

The core architecture of DiDrive is driven by two closely synergistic components: the RHDif model and the 3DICE policy optimization paradigm. Rather than operating in isolation, these two modules collaborate across the perception and decision-making stages. Specifically, RHDif serves as the foundational behavior learning module; it filters high-dimensional environmental redundancy to extract risk-aware state representations, constructing a representative multimodal behavior prior. Building upon this structural prior, 3DICE acts as the distribution correction paradigm. It leverages the representations and score functions provided by RHDif to perform in-sample guidance and stabilized optimization in the action space. Through this synergy, RHDif enables the model to effectively perceive essential local risks, while 3DICE constrains the policy updates within the data support, jointly enhancing the reliability of the autonomous driving policy without explicitly querying hazardous unseen actions. To elaborate on this framework, we first formalize the driving problem and the inherent heavy-tailed reward design in Section~\ref{sec:problem_formulation}. Building upon this formulation, we then detail the 3DICE paradigm and the RHDif architecture in the subsequent subsections.

\subsection{Problem Formulation and Reward Design}
\label{sec:problem_formulation}

We model the autonomous driving decision-making process as a standard Markov Decision Process (MDP), defined by the tuple $\mathcal{M} = \langle \mathcal{S}, \mathcal{A}, \mathcal{P}, \mathcal{R}, \gamma \rangle$. 
Here, $\mathcal{S}$ denotes the continuous state space, $\mathcal{A}$ represents the continuous action space, $\mathcal{P}(s'|s,a)$ is the unknown transition probability of the traffic environment, $\mathcal{R}(s,a)$ defines the reward function, and $\gamma \in [0, 1)$ is the discount factor. 
In our offline reinforcement learning paradigm, the agent cannot interact with the environment to collect new experiences. Instead, it must learn an optimal policy solely from a static dataset $\mathcal{D} = \{(s_i, a_i, r_i, s'_{i})\}_{i=1}^{N}$ collected by behavior policies. 

In this driving MDP, the state $s \in \mathcal{S}$ is constructed as a high-dimensional multimodal vector, encompassing the ego-vehicle kinematics, lane geometry, LiDAR point clouds, surrounding vehicle states, and navigation waypoints. The action $a \in \mathcal{A}$ comprises continuous longitudinal and lateral control commands. 

To effectively guide the offline policy optimization and ensure the safety of the autonomous driving agent, we design a comprehensive reward function $\mathcal{R}(s,a)$ that jointly encourages efficient driving, lane keeping, smooth control, and safety. Specifically, the total reward is defined as:
\begin{equation}
r = r_{speed} + r_{lane} + r_{smooth} + r_{stop} + r_{collision} + r_{\textit{offroad}},
\label{eq:reward_function}
\end{equation}
where the desired-speed tracking reward is computed as:
\begin{equation}
r_{speed} = 
\begin{cases}
v, & v \leq v_{des},\\
-(v-v_{des}), & v > v_{des},
\end{cases}
\end{equation}
where $v$ denotes the ego-vehicle speed and $v_{des}$ denotes the desired speed. The lane-keeping and smoothness terms are defined as:
\begin{equation}
r_{lane} = -|\Delta d_{lat}|,\qquad 
r_{smooth} = -0.5|a_{lat}|,
\end{equation}
where $\Delta d_{lat}$ denotes the lateral offset from the lane center and $a_{lat}$ denotes the lateral acceleration. To discourage unnecessary stopping in free-flow conditions, a stopping penalty is applied when the front space is sufficient but the ego vehicle remains nearly stationary:
\begin{equation}
r_{stop} = 
\begin{cases}
-1, & d_{front}>10.0 \ \text{and}\ v<0.1,\\
0, & \text{otherwise},
\end{cases}
\end{equation}
where $d_{front}$ denotes the front-vehicle distance. In addition, hard penalties are imposed on collision and off-road events:
\begin{align}
r_{collision} &= \begin{cases} -100, & \text{collision occurs},\\ 0, & \text{otherwise}. \end{cases} \\
r_{\textit{offroad}} &= \begin{cases} -100, & \text{off-road occurs},\\ 0, & \text{otherwise}. \end{cases}
\end{align}

This reward design physically aligns with the safety-critical nature of autonomous driving: while normal driving behaviors yield small, dense, and smooth rewards such as $r_{speed}$ and $r_{lane}$, it is essential to impose severe penalties on critical failures, setting $r_{collision}=-100$ for instance, to prevent reward hacking and ensure a reliable safety margin. Consequently, this requirement naturally introduces a highly imbalanced, heavy-tailed reward distribution. Directly applying standard offline RL or diffusion policies to such imbalanced datasets often leads to value overestimation and gradient oscillation. It is this inherent and realistic MDP property that motivates our proposed 3DICE optimization paradigm and RHDif architecture, which we detail in the following subsections.

\subsection{3DICE}

We propose the 3DICE paradigm. To address the overestimation problem caused by value functions assigning excessively high values to unseen or out-of-distribution actions, we construct a four-stage mechanism of "Representation-Guidance-Optimization-Selection." These four stages correspond to behavior prior modeling, in-sample policy guidance, numerically stable optimization, and candidate action selection, respectively. The goal is to reduce the influence of value estimation for out-of-distribution actions on policy updates and to improve the training stability of diffusion policies on heavy-tailed driving data.

\textbf{Representation}: In this stage, we use the RHDif architecture to construct a driving behavior prior. Standard diffusion policies typically perform unified conditional modeling over input states, making them susceptible to interference from background vehicles, weakly relevant road elements, and sensor noise in high-density traffic scenarios. To this end, RHDif introduces local risk awareness and global semantic filtering into diffusion-based behavior modeling: the low-level module enhances state representations of potential risk interactions through RGSE, while the high-level module suppresses background redundancy and pseudo-risk responses through CMCM. Through this risk-aware representation, RHDif can capture multimodal driving behaviors such as overtaking, yielding, car-following, and obstacle avoidance, while allowing the generation process to rely more heavily on safety-related state information. This diffusion prior further provides the action-space score function $\nabla_a \log \pi^{\mathcal{D}}(a|s)$ of the behavior policy, offering a fundamental direction for subsequent in-sample guidance. As a result, the subsequent reverse denoising is firmly grounded in a high-fidelity, risk-filtered behavioral manifold.

\textbf{Guidance}: We introduce the idea of distribution correction estimation and perform in-sample reweighting of the behavior policy through a density ratio. In distribution correction theory, the stationary distribution correction ratio between the target policy and the behavior distribution has been used to alleviate distribution shift and value overestimation in offline policy optimization \cite{nachum2019dualdice,zhang2020gendice,lee2021optidice}. The density ratio between the optimal stationary distribution and the behavior distribution,
$w^{*}(s,a)=\frac{d^{*}(s,a)}{d^{\mathcal{D}}(s,a)}$,
characterizes the relative importance of data samples under the target policy. From the normalization relationship between the state-action stationary distribution and the conditional policy, the target policy can be expressed as a weighted form of the behavior policy, namely $\pi^{*}(a|s) \propto w^{*}(s,a)\pi^{\mathcal{D}}(a|s)$.

This relationship indicates that policy improvement can be achieved by increasing the generation probability of actions corresponding to high-weight samples, without directly searching for optimal actions in unsupported regions of the action space. Therefore, we first use a diffusion model to learn the behavior prior $\pi^{\mathcal{D}}$, and then introduce a guidance term induced by the density ratio during the reverse sampling process. Specifically, the score functions of the target policy and the behavior policy at diffusion timestep $t$ satisfy the following relationship:
\begin{equation}
\begin{split}
\nabla_{a_{t}}\log\pi_{t}^{*}(a_{t}|s) = & \, \nabla_{a_{t}}\log\pi_{t}^{\mathcal{D}}(a_{t}|s) \\
& + \nabla_{a_{t}}\log \mathbb{E}_{a_{0}\sim\pi^{\mathcal{D}}(a_{0}|a_{t},s)} \left[w^{*}(s,a_{0})\right].
\end{split}
\label{eq:score_guidance}
\end{equation}

Here, the first term corresponds to the diffusion score of the behavior prior, while the second term represents the policy correction direction induced by the density ratio. Since the second term contains a conditional expectation, it is difficult to compute directly. Therefore, we introduce a neural network
$g_{\theta}(s,a_t,t)$ to approximate
$\log\mathbb{E}_{a_0\sim\pi^{\mathcal{D}}(a_0|a_t,s)}[w^{*}(s,a_0)]$,
and propose the following In-sample Guidance Learning (IGL) objective:
\begin{equation}
\begin{split}
\min_{\theta} \mathbb{E}_{t\sim\mathcal{U}(0,T)} & \, \mathbb{E}_{a_0\sim\pi^{\mathcal{D}}(a_0|s)} \mathbb{E}_{a_t\sim p(a_t|a_0)} \\
& \cdot \left[ w^{*}(s,a_0)e^{-g_{\theta}(s,a_t,t)} + g_{\theta}(s,a_t,t) \right].
\end{split}
\label{eq:igl}
\end{equation}

Under this objective, $\nabla_{a_t}g_{\theta}(s,a_t,t)$ can be used to approximate the policy correction term in Eq.~\eqref{eq:score_guidance}. Since the training samples are drawn from offline data and its diffusion forward perturbation process, IGL does not require the additional generation or evaluation of arbitrary actions, seamlessly constraining the policy optimization strictly within the data support.

In the dual formulation of the DICE objective, the optimal density ratio is derived from the advantage function. Consequently, when implementing IGL, it is crucial to ensure that the mapping $w^{*}(s,a)=(f')^{-1}(Q^{*}(s,a)-V^{*}(s))$ remains strictly non-negative. To this end, we adopt a piecewise $f$-divergence: a quadratic constraint is used when the weight falls into the high-weight region, while an entropy-based constraint is applied when the weight falls into the low-weight region. This design ensures that the weight mapping $(f')^{-1}(\cdot)$ remains non-negative across different intervals, thereby satisfying the basic requirement of IGL for sample weights. Meanwhile, the corresponding convex conjugate function $f^{*}$ grows at most quadratically in the positive interval. Compared with exponential growth, this weakens the amplification effect of extreme advantage values on DICE value updates, thereby improving optimization stability in heavy-tailed reward scenarios.

\textbf{Optimization}: With the model architecture and guidance direction established in the preceding stages, this third stage focuses on stabilizing the training process against data imbalance. In autonomous driving, safety-critical edge cases are extremely sparse and usually appear as heavy-tailed reward signals. Most samples correspond to normal driving with mild penalties, whereas a small number of collision-related samples may produce large negative returns. Directly optimizing diffusion policies on such imbalanced data can destabilize value estimation, amplify estimation bias, and trigger gradient oscillation, leading to distorted guidance signals and unstable control outputs.

To stabilize training, we propose a spatiotemporal collaborative optimization mechanism. In the spatial dimension, we regularize value estimation and the guidance field to suppress extreme estimation errors. Specifically, for the value network, we introduce an error clipping operation $\text{clamp}(\cdot,0,c_{\text{clip}})$ to bound the influence of extreme samples. We further add a value regularization term $\lambda_v\tilde{V}(s)^2$ to constrain extrapolated values for states outside the data support, preventing abnormal overestimation.

For the energy function $g_{\theta}$ used to guide diffusion sampling, numerical oscillations may destabilize the generated actions. Therefore, we introduce statistical regularization to smooth the guidance field. A variance regularization term $\lambda_{\text{var}}\mathrm{Var}(\tilde{g}_{\theta})$ suppresses high-frequency fluctuations in the energy surface, while a mean regularization term $\lambda_{\text{mean}}|\mathbb{E}[\tilde{g}_{\theta}]|$ anchors the guidance signal around zero and prevents systematic numerical drift.

In the temporal dimension, mini-batch sampling may introduce high-frequency noise across training iterations. To smooth the parameter update trajectory, we introduce Progressive Parameter Integration (PPI), which acts as a temporal low-pass filter during optimization. By computing an exponential moving average of historical network states, PPI mitigates the disproportionate impact of individual extreme mini-batches, thereby dampening high-frequency gradient oscillations. Rather than relying solely on instantaneous estimates, PPI maintains smoothly evolving target estimates:
\begin{align}
\tilde{V}^{(k)}(s) &= \eta \cdot \tilde{V}^{(k-1)}(s) + (1-\eta) \cdot V_{\theta}^{(k)}(s), \\
\tilde{g}_{\theta}^{(k)}(s,a_{\tau},\tau) &= \eta \cdot \tilde{g}_{\theta}^{(k-1)}(s,a_{\tau},\tau) \notag \\
&\quad + (1-\eta) \cdot g_{\theta}^{(k)}(s,a_{\tau},\tau),
\end{align}
where $k$ denotes the training iteration, $\tau$ denotes the diffusion timestep, $\eta \in [0,1)$ is the decay coefficient, and $V_{\theta}^{(k)}$ and $g_{\theta}^{(k)}$ denote the instantaneous state-value estimate and guidance-term estimate of the current network, respectively.

Based on the above mechanism, \textbf{the value-side objective} is defined as:
\begin{equation}
\begin{aligned}
\mathcal{L}_{\text{value}}(\theta) = 
&\alpha \mathbb{E}_{d^{\mathcal{D}}} \left[
\text{clamp}\left(
f^{*}\left( \frac{\tilde{Q}-V_{\theta}}{\alpha} \right),
0,
c_{\text{clip}}
\right)
\right] \\
&+\mathbb{E}_{d^{\mathcal{D}}}\left[(1-\gamma)V_{\theta}(s)\right]
+\lambda_v \mathbb{E}_{d^{\mathcal{D}}}\left[V_{\theta}(s)^2\right],
\label{eq:value_loss_smooth}
\end{aligned}
\end{equation}
where $d^{\mathcal{D}}$ denotes the empirical distribution of the offline dataset, $\gamma$ is the discount factor, $\alpha$ controls the strength of the dual constraint, $c_{\text{clip}}$ is the clipping threshold, and $\lambda_v$ is the value regularization coefficient.

Similarly, \textbf{the energy-side objective} is defined as:
\begin{equation}
\mathcal{L}_{\text{dual}} =
\mathbb{E}_{\pi^{\mathcal{D}},p(a_{\tau}|a_0)}
\left[
w \cdot e^{-\tilde{g}}+\tilde{g}
\right]
+
\lambda_{\text{var}}\mathrm{Var}(\tilde{g})
+
\lambda_{\text{mean}}|\mathbb{E}[\tilde{g}]|,
\label{eq:dual_loss_smooth}
\end{equation}
where $\mathbb{E}_{\pi^{\mathcal{D}},p(a_{\tau}|a_0)}$ denotes the joint expectation over the behavior policy and the diffusion transition probability, $w$ is the importance weight derived from value estimation, and $\lambda_{\text{var}}$ and $\lambda_{\text{mean}}$ control the variance and magnitude of the guidance signal, respectively.

\textbf{Selection}: As the final stage of the RGOS mechanism, this step identifies a high-quality action from the diffusion-generated candidates while ensuring the final decision remains tightly anchored to the behavior-supported action region.

First, we use the diffusion policy $\pi_{\theta}$ trained in the representation stage as the generator. Given the current state $s$, we sample in parallel a set of candidate actions $\mathcal{A}_{raw} = \{a^i\}_{i=1}^K$ that are consistent in intention but differ in detail, where $K$ denotes the number of parallel samples. To avoid the high-variance risk caused by a single value estimate, we introduce the critics $\{Q_{\phi_j}^*\}_{j=1}^M$, which are stably trained through the DICE objective in the Optimization stage, as evaluators. Unlike traditional Softmax sampling, which inherently assigns non-zero probabilities to sub-optimal actions and may expose the agent to stochastic risks, we adopt an ensemble-based rejection sampling strategy. This approach effectively filters out lower-confidence candidates and selects the action with the highest consensus value, providing a tighter safety margin for the final decision:
\begin{equation}
a_{stat}^* = \arg\max_{a^i \in \mathcal{A}_{raw}} \left( \frac{1}{M} \sum_{j=1}^{M} Q_{\phi_j}^*(s, a^i) \right),
\end{equation}
where $a_{stat}^*$ denotes the optimal action, $\mathcal{A}_{raw}$ represents the sampled set containing multiple candidate actions, and $M$ denotes the number of ensemble Q-networks.

Since the candidate actions are generated by a diffusion policy fitted to behavioral data, they are generally closer to the high-probability regions covered by the behavior data than actions obtained through direct optimization in the continuous action space \cite{wu2022spot}. The critics perform relative ranking within this candidate set rather than conducting a global search over arbitrary unsupported actions, thereby effectively reducing the influence of out-of-distribution value overestimation on action selection.

\subsection{Risk-Aware Hierarchical Diffusion}
\par\vspace{1em}

\begin{center}
\begin{minipage}{0.95\columnwidth}
    \centering
    \includegraphics[width=\linewidth]{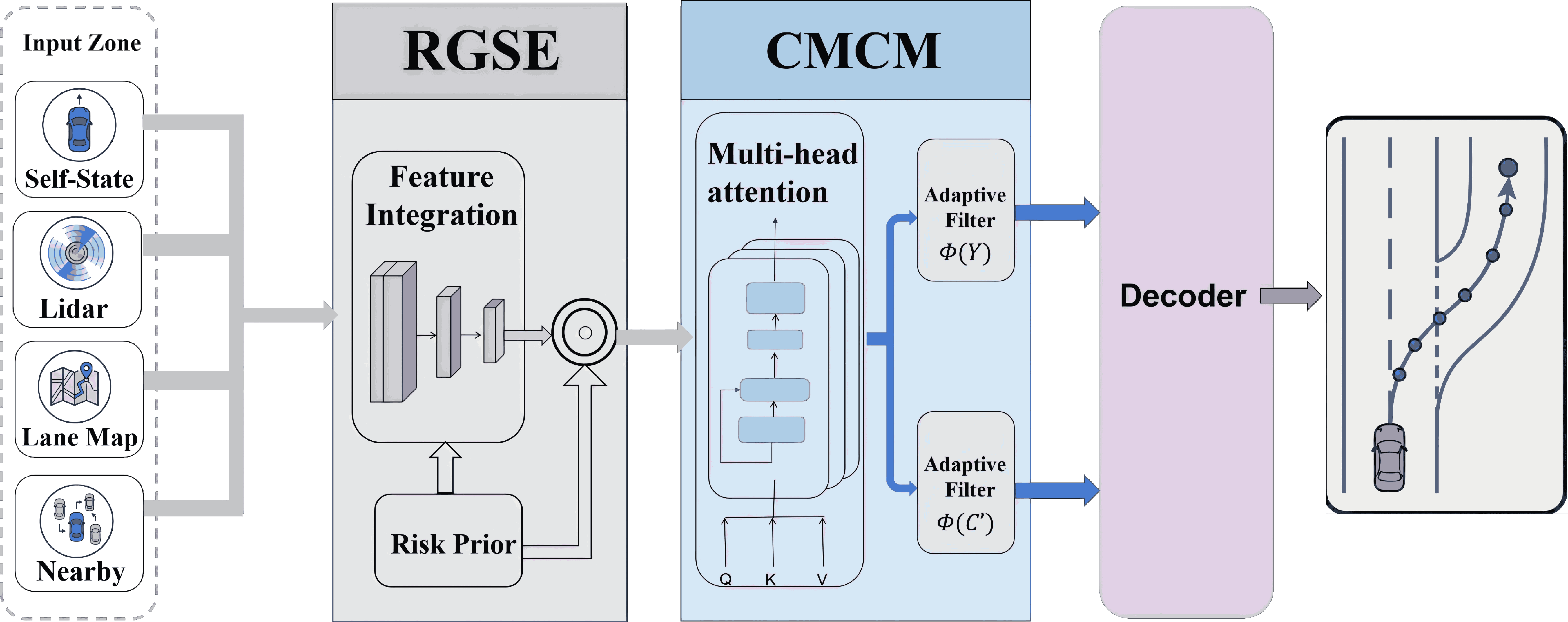}
    \captionof{figure}{Overview of the RHDif architecture. RGSE enhances local risk-aware features, while CMCM calibrates risk signals with global semantic context.}
    \label{fig:hierarchical_diffusion}
\end{minipage}
\end{center}

To address environmental information redundancy and the difficulty of distinguishing local risk factors in complex traffic scenarios, this paper proposes RHDif, a risk-aware spatiotemporally decoupled hierarchical diffusion architecture, as shown in Fig.~\ref{fig:hierarchical_diffusion}. This architecture performs hierarchical feature selection in the latent space to balance rapid hazard perception with global scene comprehension. At the low level, the RGSE enhances local safety-related features to prevent critical threats from being diluted by background noise. At the high level, the CMCM incorporates global semantic information to calibrate these local risk responses, effectively preventing overly conservative or erratic behaviors triggered by pseudo-risks.

Specifically, RGSE generates risk-related weights based on kinematic priors—such as the relative positions of neighboring vehicles, longitudinal distance, lateral offset, and relative velocity—enabling the model to assign stronger feature responses to regions with elevated potential risk. Concurrently, CMCM models the cross-modal relationship between trajectory intentions and environmental context during the diffusion denoising process, suppressing redundant background features through channel and spatial gating. Through this hierarchical mechanism, proceeding from local risk enhancement to global semantic calibration, RHDif precisely allocates its representational capacity to the most safety-critical interactions, thereby significantly reducing the interference of high-dimensional redundant information in action generation.

\subsubsection{Risk-Gated Spatiotemporal Encoder Module}

In complex traffic flows, surrounding vehicles, lane structures, and sensor observations introduce substantial redundant information. If all neighboring objects are assigned uniform importance, the safety-critical cues can be easily overwhelmed by background noise. To address this issue, RGSE first employs local risk awareness based on kinematic priors to model the local interactions between surrounding vehicles and the ego vehicle. Let the raw observation of the $i$-th surrounding vehicle be $o_i \in \mathbb{R}^4$. Its geometric importance weight is defined as:
\begin{equation}
\mathcal{M}_{geo}^{(i)} = \sigma \left( \frac{\tau_{long} - x^{(i)}}{\kappa_1} \right) \cdot \sigma \left( \frac{\tau_{lat} - |y^{(i)}|}{\kappa_2} \right) \cdot \mathbb{I}(x^{(i)} > 0),
\label{eq:geo_gate}
\end{equation}
where $\mathcal{M}_{geo}^{(i)} \in (0, 1)$ denotes the spatial importance of the $i$-th surrounding vehicle, $\tau_{long}$ and $\tau_{lat}$ denote the longitudinal perception range and lateral attention range, respectively, $x^{(i)}$ and $y^{(i)}$ are the local coordinates of the surrounding vehicle relative to the ego vehicle, $\kappa_1$ and $\kappa_2$ control the smoothness of weight variation, and $\mathbb{I}(\cdot)$ is the indicator function used to retain the region in front of the ego vehicle, where $x^{(i)} > 0$. This weight is then applied to the projected features of surrounding vehicles, and the features of all surrounding vehicles are weighted and aggregated to obtain the local vehicle interaction representation $\mathbf{h}_{veh}$.

After multimodal information fusion, we concatenate the ego state, lane information, LiDAR features, and surrounding-vehicle interaction representation, and add learnable type embeddings to form the multimodal context tensor
$\mathbf{Z}_{ctx} = \text{Concat}(\mathbf{h}_{ego}, \mathbf{h}_{lane}, \mathbf{h}_{lidar}, \mathbf{h}_{veh}) + \mathbf{E}_{type}$.
To further enhance the model’s sensitivity to local risk variations, RGSE introduces a risk-gating mechanism that transforms spatiotemporal risk priors into channel-level modulation weights. Specifically, we construct the risk input vector
$\mathbf{r}_{risk} = [v, |\Delta d_{lat}|, d_{front}, \Delta v]$,
and generate dynamic risk weights through a multilayer perceptron:
\begin{equation}
\mathbf{W}_{risk} = \sigma(\mathcal{F}_{mlp}(\mathbf{r}_{risk})) + c_{base},
\label{eq:risk_weight}
\end{equation}
where $c_{base}$ denotes the base activation level, which prevents the risk weights from becoming too small. Then, $\mathbf{W}_{risk}$ is applied to the multimodal context features as a channel-level multiplicative modulation factor, yielding
$\tilde{\mathbf{Z}}_{ctx} = \mathbf{Z}_{ctx} \odot \mathbf{W}_{risk}$.
When the relative velocity increases, the longitudinal distance decreases, or the lateral offset increases, the risk-gating mechanism increases the response weights of the relevant modalities and channels, enabling the diffusion policy to more fully utilize safety-related state information during reverse denoising. Compared with fixed-threshold rules, this mechanism modulates feature responses in a continuous and differentiable manner, helping alleviate the discontinuities caused by hard thresholds.

\subsubsection{Cross-Modal Contextual Modulator Module}

In dynamic traffic scenarios, local risk signals are not always equivalent to real hazards. For example, vehicles in the opposite lane, distant static obstacles, or transient sensor noise may produce strong responses at the local feature level, but they do not necessarily require conservative actions. Therefore, relying solely on low-level risk enhancement may lead to misjudgment or overly conservative behavior. To alleviate this issue, CMCM introduces cross-modal contextual modulation in the high-level latent space, using global semantic information to filter and calibrate local risk responses.

First, we formulate the risk-modulated context tensor $\tilde{\mathbf{Z}}_{ctx}$ from the RGSE module as a sequence of condition tokens $C \in \mathbb{R}^{B \times N \times D}$. A multi-head self-attention mechanism is then applied to model global dependencies among these tokens, yielding the contextualized representation $C'$. Subsequently, let $Y$ denote the noisy action latent at the current diffusion timestep. For both the latent feature $Y$ and the context feature $C'$, CMCM employs a unified channel-spatial gating operator:
\begin{equation}
    \Phi(X)=X \odot \sigma(\mathcal{M}_{channel}(X)) 
    \odot \sigma(\mathcal{M}_{spatial}(\text{Pool}(X))),
\label{eq:cmcm_gate}
\end{equation}
where $X$ may denote either the latent feature $Y$ or the context feature $C'$, and $\mathcal{M}_{channel}$ and $\mathcal{M}_{spatial}$ denote the channel attention block and the spatial attention block, respectively. This gating operator enhances features related to interaction-aware decision-making while suppressing redundant channel responses and background regions.

Finally, the modulated context feature $\Phi(C')$ and latent state feature $\Phi(Y)$ are injected into the diffusion denoising backbone through residual fusion:
\begin{equation}
    Z_{out} = \mathcal{W}_{fuse}(\Phi(C')) + \mathcal{W}_{self}(\Phi(Y)) + Y_{residual},
\label{eq:cmcm_fusion}
\end{equation}
where $\mathcal{W}_{fuse}$ and $\mathcal{W}_{self}$ denote linear projection layers. Through this cross-modal contextual modulation mechanism, CMCM incorporates global scene constraints while preserving local risk responses, thereby reducing the interference of irrelevant high-response features. Coupled with the RGSE, this hierarchical gating directly injects risk-calibrated, high-fidelity priors into the diffusion denoising backbone.

\section{Experiments}

\subsection{Experimental Setup}

To validate the effectiveness of the proposed algorithm under high-dimensional states and dynamic interactions, we construct a closed-loop experimental platform using the high-fidelity autonomous driving simulator CARLA \cite{dosovitskiy2017carla}. Specifically, we select Town03, a complex urban environment featuring a multi-lane roundabout, tunnels, and varied irregular junctions, for model training and in-domain evaluation. To assess zero-shot cross-town generalization, we introduce Town05, an unseen map characterized by grid-like multi-lane networks and highway intersections. The layouts of both maps are illustrated in Fig.~\ref{fig:town_maps}. Furthermore, to comprehensively evaluate the representational robustness of the model, we set up simple, medium, and complex scenarios with 20, 40, and 60 background vehicles, respectively, on both maps.

\par\vspace{1em}
\begin{center}
\begin{minipage}{0.95\columnwidth}
    \centering

    \begin{minipage}[b]{0.49\linewidth}
        \centering
        \includegraphics[width=\linewidth]{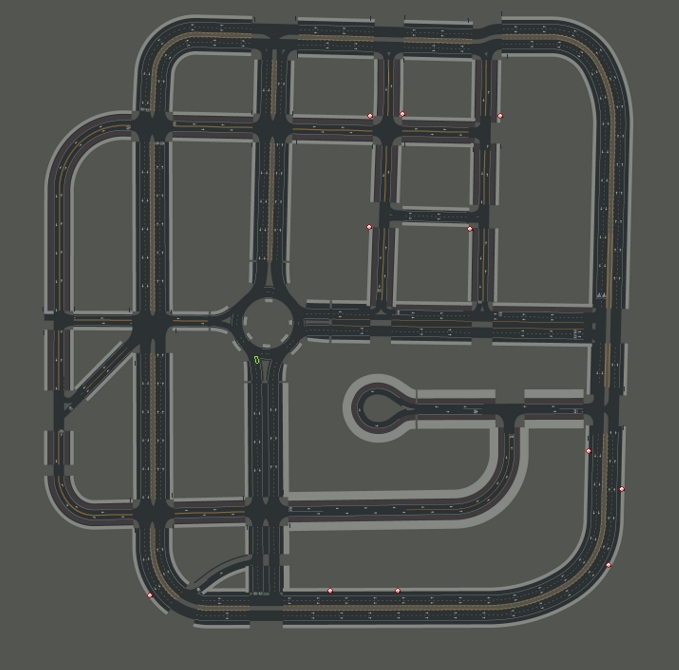}
        
        \small (a) Town03
    \end{minipage}
    \hfill
    \begin{minipage}[b]{0.49\linewidth}
        \centering
        \includegraphics[width=\linewidth]{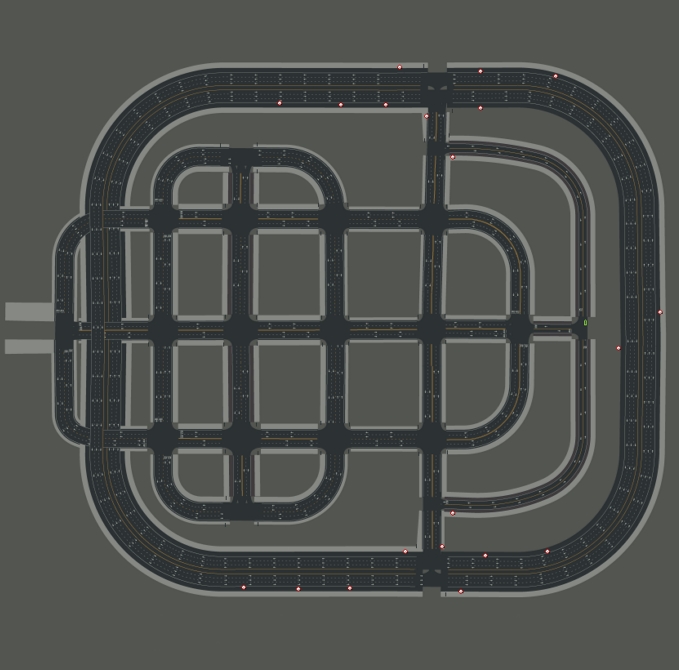}
        
        \small (b) Town05
    \end{minipage}

    \captionof{figure}{Top-view maps of CARLA Town03 and Town05.}
    \label{fig:town_maps}
\end{minipage}
\end{center}

The experiments use a large-scale offline dataset stored in HDF5 format. This dataset is collected in Town03 and contains more than 7,000 trajectories and 1.1 million timesteps. The data distribution consists of a mixture of expert policies and random exploration policies at a ratio of $8:2$. The model input is defined as a 307-dimensional observation vector, including the ego vehicle state ($9$ dimensions), lane information ($2$ dimensions), LiDAR point-cloud features ($240$ dimensions), local surrounding-vehicle features ($20$ dimensions), and navigation waypoints ($36$ dimensions).

The training process adopts a two-stage strategy. In the first stage, the backbone of the RHDif architecture is pretrained with a learning rate of $3 \times 10^{-4}$ and a batch size of $1024$ to extract a high-fidelity behavior prior. In the second stage, the 3DICE paradigm is fine-tuned for distribution correction. In the experiments, the number of denoising iterations of the diffusion model $K$ is set to 15, the inference sampling number \texttt{inference\_sample} is set to 32, and the number of ensemble estimation samples $M$ is set to 16. Additionally, the learning rate of the value network is set to $4 \times 10^{-5}$, and the discount factor $\gamma$ is set to 0.99 to maintain the stability of long-horizon optimization.

During evaluation, we conduct 100 online closed-loop test episodes for each configuration, using Success Rate (SR), Average Reward, and Average Distance as the core evaluation metrics\cite{Jia_2024_NeurIPS_Bench2Drive,Caesar2020nuScenes,Ettinger2021WOMD}. SR is defined as the proportion of test episodes in which the vehicle neither collides nor drives off the road within 1000 timesteps. Average Reward represents the cumulative task returns obtained per episode, averaged across all test runs, which comprehensively reflects the overall optimality of the policy. Average Distance measures the mean longitudinal distance traveled by the ego vehicle before an episode terminates due to success, failure, or timeout, serving as a quantitative indicator of driving progress and survival capability.

For a fair comparison, all baseline methods—including Implicit Q-Learning (IQL) \cite{Kostrikov2021OfflineRL}, Conservative Q-Learning (CQL) \cite{NEURIPS2020_0d2b2061}, OptiDICE (ODICE) \cite{lee2021optidice}, TD3-BC \cite{Fujimoto2021TD3BC}, Q-Guided Policy Optimization (QGPO) \cite{10.5555/3618408.3619356}, and Diffusion-QL \cite{Wang2022DiffusionPA}—are trained and evaluated under an identical experimental protocol. Specifically, they share the same offline dataset, state-action space representations, reward function, and closed-loop evaluation configurations. Furthermore, diffusion-based baselines (QGPO and Diffusion-QL) adopt the exact same inference parameters, such as the number of denoising steps and sampled candidates, as DiDrive. By strictly controlling these variables, we ensure that any performance disparities stem directly from the underlying policy learning and optimization mechanisms.

\subsection{Comparative Experiments}
\subsubsection{Training Analysis}

To evaluate the learning efficiency and stability of our proposed DiDrive framework, we first analyze the convergence dynamics of different baseline methods during the training phase.

\begin{center}
\begin{minipage}{0.95\columnwidth}
    \centering
    \includegraphics[width=\linewidth]{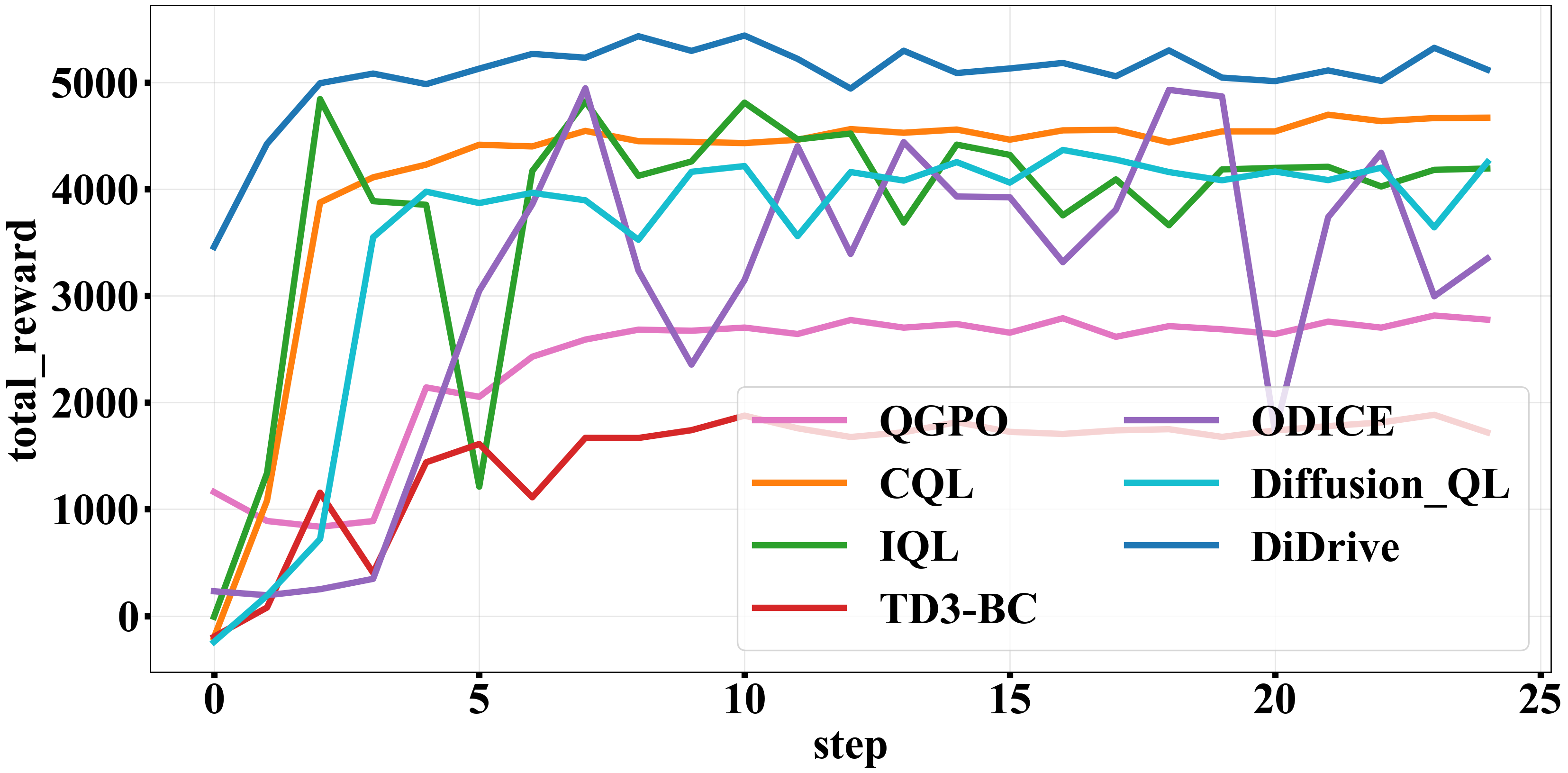}
    \captionof{figure}{Learning curves of total reward for different methods during training.}
    \label{fig:total_reward}
\end{minipage}
\end{center}

\begin{center}
\begin{minipage}{0.95\columnwidth}
    \centering
    \includegraphics[width=\linewidth]{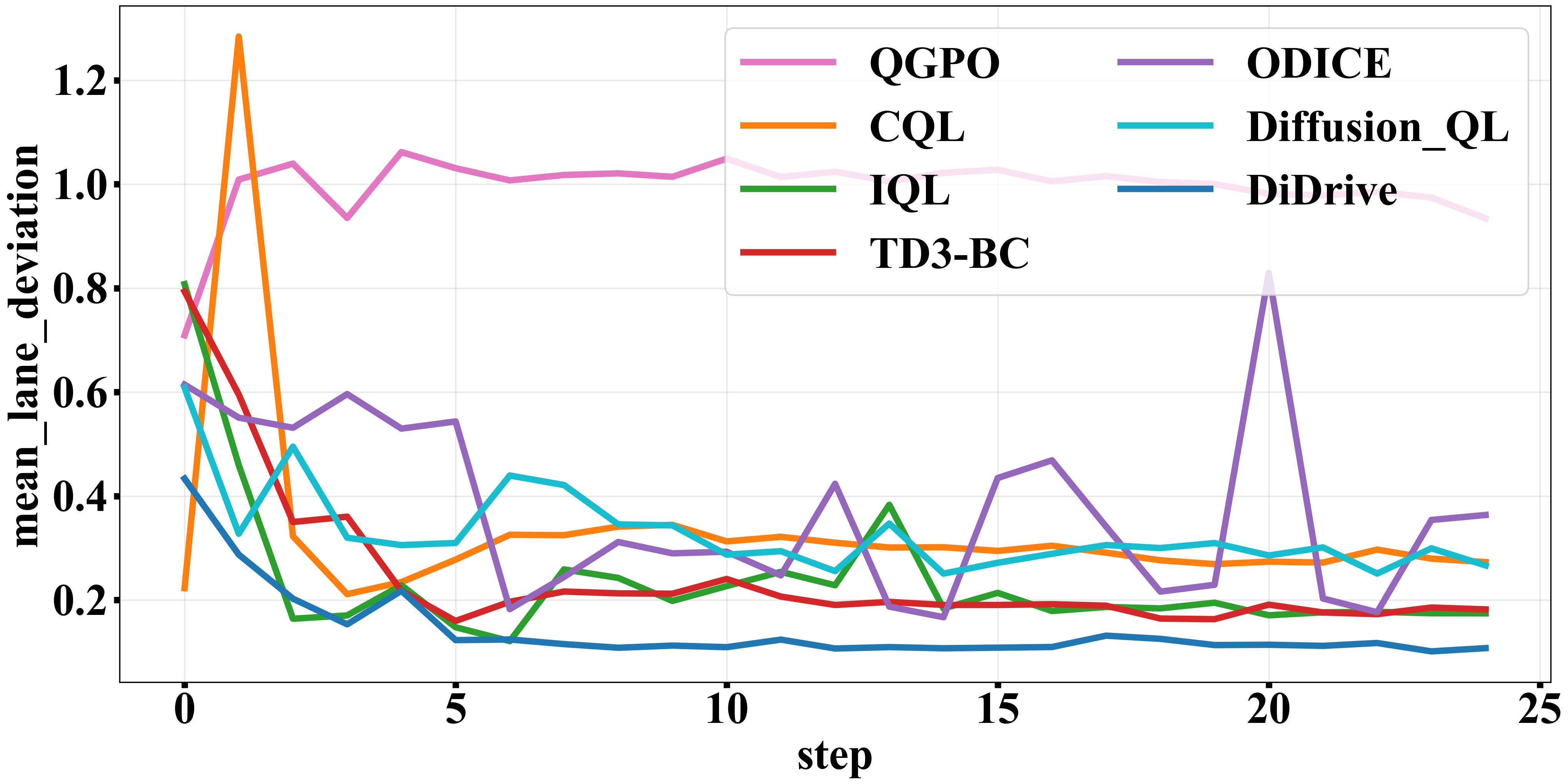}
    \captionof{figure}{Learning curves of lane deviation for different methods during training.}
    \label{fig:road_deviation}
\end{minipage}
\end{center}

Figs.~\ref{fig:total_reward} and~\ref{fig:road_deviation} illustrate the overall convergence dynamics of DiDrive and the baseline methods during training. Overall, DiDrive shows superior performance and stability, while the baseline methods exhibit different limitations when handling high-dimensional complex states. The diffusion-based baseline QGPO shows long-term stagnation in total reward, while its lane deviation remains above 1.0 m for most of the training process. This indicates that, without sufficiently reliable value guidance, standard diffusion policies are prone to extrapolation errors in large state spaces and struggle to maintain lateral control constraints. Although TD3-BC keeps the deviation at a relatively low level, its low reward suggests that the policy suffers from unstable speed control in dynamic interactions. Specifically, it exhibits severe frequent oscillations, alternating between excessive acceleration and extremely low-speed stagnation, which limits its overall driving efficiency and long-horizon stability.

The strongly oscillating curve of ODICE reflects the numerical instability of traditional DICE-based methods when facing high-dimensional features and heavy-tailed rewards. The reward degradation and deviation increase in the later stage of training further indicate potential value divergence caused by unstable gradients. In comparison, IQL and CQL show a certain degree of competitiveness. However, the relatively large performance fluctuations of IQL suggest that its expectile regression mechanism may lack sufficient consistency in long-horizon decision-making, while the conservative nature of CQL limits its ability to approach the optimal performance boundary. In contrast, benefiting from the high-fidelity prior features provided by the hierarchical diffusion architecture and the smooth optimization introduced by the PPI mechanism in the 3DICE paradigm, DiDrive rapidly converges to the highest reward range ($>5200$) within a small number of training steps, while maintaining the lane deviation at an extremely low level of approximately 0.1 m throughout training. This smooth and efficient convergence trend demonstrates that DiDrive effectively mitigates the interference of feature redundancy and environmental noise, achieving a favorable balance between traffic efficiency and control precision.

\subsubsection{Testing Analysis}

Following the training phase, we comprehensively evaluate the closed-loop driving performance and cross-domain generalization capabilities of all fully trained models under varying traffic densities.

\begin{table*}[t]
    \centering
    \caption{Performance comparison on Town03 under different traffic densities. We report Success Rate, Average Reward, and Average Driving Distance over 100 closed-loop evaluation episodes.}
    \label{tab:town03_benchmark}
    \resizebox{\textwidth}{!}{
    \setlength{\tabcolsep}{4pt}
    \renewcommand{\arraystretch}{1.2}
    \begin{tabular}{llccccccc}
        \toprule
        \multirow{2}{*}{\textbf{Scenario}} & \multirow{2}{*}{\textbf{Metric}} 
        & \multicolumn{7}{c}{\textbf{Method}} \\
        \cmidrule(lr){3-9}
        & & IQL~\cite{Kostrikov2021OfflineRL} & CQL~\cite{NEURIPS2020_0d2b2061} & ODICE~\cite{lee2021optidice} & TD3-BC~\cite{Fujimoto2021TD3BC} & QGPO~\cite{10.5555/3618408.3619356} & Diffusion-QL~\cite{Wang2022DiffusionPA} & \textbf{DiDrive (Ours)} \\
        \midrule
        \multirow{3}{*}{\shortstack[l]{Simple\\20 Vehicles}}
        & Success Rate $\uparrow$ & 77\% & 86\% & 21\% & 87\% & 16\% & 90\% & \textbf{92\%} \\
        & Avg. Reward $\uparrow$ & 3943.99 & 4004.43 & 1174.92 & 955.72 & 1157.62 & 3978.19 & \textbf{4902.94} \\
        & Avg. Distance (m) $\uparrow$ & 306 & 247 & 171 & \textbf{327} & 125 & 241 & 286 \\
        \midrule
        \multirow{3}{*}{\shortstack[l]{Medium\\40 Vehicles}}
        & Success Rate $\uparrow$ & 56\% & 83\% & 20\% & 83\% & 14\% & 85\% & \textbf{88\%} \\
        & Avg. Reward $\uparrow$ & 3488.71 & 3922.53 & 1063.85 & 787.72 & 1148.52 & 3823.84 & \textbf{4713.84} \\
        & Avg. Distance (m) $\uparrow$ & 269 & \textbf{280} & 125 & 276 & 122 & 236 & 276 \\
        \midrule
        \multirow{3}{*}{\shortstack[l]{Complex\\60 Vehicles}}
        & Success Rate $\uparrow$ & 33\% & 74\% & 17\% & 72\% & 9\% & 77\% & \textbf{85\%} \\
        & Avg. Reward $\uparrow$ & 3275.79 & 3819.05 & 1017.21 & 539.95 & 927.71 & 3425.91 & \textbf{4295.68} \\
        & Avg. Distance (m) $\uparrow$ & 237 & 245 & 113 & 238 & 103 & 222 & \textbf{254} \\
        \bottomrule
    \end{tabular}
    }
\end{table*}

Table~\ref{tab:town03_benchmark} quantifies the closed-loop performance of different methods under varying traffic densities. In the complex scenario with 60 background vehicles, traditional baselines reveal clear performance bottlenecks. CQL maintains a success rate of 74.00\% through conservative value constraints, but its average reward is only 3819.05, which is lower than DiDrive's 4295.68. This suggests that excessive conservatism, while helping avoid some high-risk actions, may also limit driving efficiency and introduce unnecessary low-speed behaviors, which are penalized by the desired-speed tracking and stopping terms in the reward function. IQL remains somewhat competitive in simple scenarios, but its success rate drops to 33.00\% in complex scenarios, indicating that its implicit policy improvement mechanism lacks stability under high-density traffic flows and long-horizon interactions. Although TD3-BC with behavior cloning regularization achieves the longest driving distance in the simple scenario (327 m), its speed control exhibits severe frequent oscillations in dynamic interactions, alternating between excessive acceleration and near-stationary behavior. This control instability substantially reduces its average reward to 539.95 in the complex scenario, because excessive acceleration, near-stationary behavior, and unstable lateral motion are directly penalized by the desired-speed tracking, stopping, and lateral-acceleration terms in the reward function.

Notably, both the purely generative policy QGPO and the distribution correction method ODICE show significant performance degradation. Across the three traffic densities, the highest success rates of QGPO and ODICE are only 16.00\% and 21.00\%, respectively. This indicates that, without stable support-set constraints or reliable density-ratio estimation, the policy is vulnerable to the combined effects of high-dimensional observations, accumulated closed-loop errors, and heavy-tailed rewards. As a representative value-guided diffusion policy, Diffusion-QL demonstrates strong multimodal expressiveness, achieving success rates of 90.00\%, 85.00\%, and 77.00\% in simple, medium, and complex scenarios, respectively. However, its average reward in the complex scenario is 3425.91, which remains significantly lower than DiDrive's 4295.68. This suggests that relying solely on the generative capability of diffusion policies and value guidance is still insufficient to fully address risk identification, background redundancy, and value estimation near the boundary of the data support in high-density traffic.

In contrast, DiDrive achieves the highest success rate and average reward across all three scenarios. In the complex scenario with 60 vehicles, DiDrive still maintains a success rate of 85.00\%, an average reward of 4295.68, and the longest average driving distance of 254 m. These results show that the advantage of DiDrive does not stem from a single metric, but is reflected in a more stable balance among safety, traffic efficiency, and trajectory quality. On the one hand, 3DICE reduces the influence of out-of-distribution value overestimation and heavy-tailed sample perturbations on policy updates through in-sample guidance, stabilized optimization, and support-set candidate selection. On the other hand, RHDif enables the model to focus more on key interactions that affect safety boundaries in high-dimensional traffic inputs through risk-aware hierarchical representation. By combining these two components, DiDrive maintains strong closed-loop robustness even as traffic density increases.

\begin{table*}[t]
    \centering
    \caption{Generalization performance on Town05 under different traffic densities. We report Success Rate, Average Reward, and Average Driving Distance over 100 closed-loop evaluation episodes.}
    \label{tab:town05_generalization}
    \resizebox{\textwidth}{!}{
    \setlength{\tabcolsep}{4pt}
    \renewcommand{\arraystretch}{1.2}
    \begin{tabular}{llccccccc}
        \toprule
        \multirow{2}{*}{\textbf{Scenario}} & \multirow{2}{*}{\textbf{Metric}} 
        & \multicolumn{7}{c}{\textbf{Method}} \\
        \cmidrule(lr){3-9}
        & & IQL & CQL & ODICE & TD3-BC & Diffusion-QL & QGPO & \textbf{DiDrive (Ours)} \\
        \midrule
        \multirow{3}{*}{\shortstack[l]{Simple\\20 Vehicles}}
        & Success Rate $\uparrow$ & 72\% & 85\% & 8\% & 81\% & 82\% & 72\% & \textbf{89\%} \\
        & Avg. Reward $\uparrow$ & 2877.13 & 2183.61 & 1116.17 & 3971.17 & 3393.62 & -22.70 & \textbf{4397.61} \\
        & Avg. Distance (m) $\uparrow$ & 299 & 188 & 102 & \textbf{336} & 218 & 55 & 306 \\
        \midrule
        \multirow{3}{*}{\shortstack[l]{Medium\\40 Vehicles}}
        & Success Rate $\uparrow$ & 68\% & 81\% & 7\% & 65\% & 77\% & 63\% & \textbf{86\%} \\
        & Avg. Reward $\uparrow$ & 2869.62 & 2078.35 & 882.48 & 3570.02 & 3311.99 & 88.80 & \textbf{4257.76} \\
        & Avg. Distance (m) $\uparrow$ & \textbf{304} & 179 & 88 & 301 & 216 & 61 & 299 \\
        \midrule
        \multirow{3}{*}{\shortstack[l]{Complex\\60 Vehicles}}
        & Success Rate $\uparrow$ & 58\% & 70\% & 7\% & 35\% & 66\% & 49\% & \textbf{81\%} \\
        & Avg. Reward $\uparrow$ & 2690.99 & 1611.80 & 502.23 & 2715.14 & 3011.24 & 103.22 & \textbf{4085.39} \\
        & Avg. Distance (m) $\uparrow$ & 264 & 145 & 78 & 251 & 199 & 58 & \textbf{283} \\
        \bottomrule
    \end{tabular}
    }
\end{table*}

To further evaluate the cross-town generalization ability of the model, we conduct additional closed-loop tests on Town05. Compared to Town03, Town05 exhibits significantly different road layouts. Therefore, testing on this map rigorously verifies whether the policy can effectively transfer to new road layouts and interaction patterns, rather than merely overfitting to the training environment \cite{Li_2024_CVPR_EgoStatus}. As shown in Table~\ref{tab:town05_generalization}, DiDrive achieves success rates of 89.00\%, 86.00\%, and 81.00\% under simple, medium, and complex traffic densities, respectively, with corresponding average rewards of 4397.61, 4257.76, and 4085.39. These results indicate that DiDrive maintains strong closed-loop performance under unseen town environments.

As traffic density increases, the baseline methods exhibit varying degrees of performance degradation on Town05. TD3-BC shows certain competitiveness in the simple scenario, but its success rate drops to 35.00\% in the complex scenario. This indicates that policies relying on behavior cloning constraints are prone to insufficient generalization when road environments and traffic interactions change. CQL still maintains a relatively high success rate on Town05, reaching 70.00\% in the complex scenario, but its average reward is only 1611.80. This suggests that the inherent conservatism of CQL is further exacerbated in unseen environments. While safe, the policy becomes overly timid and struggles to maintain driving efficiency when faced with unfamiliar road topologies.

Diffusion-QL achieves a success rate of 66.00\% and an average reward of 3011.24 in the complex scenario, outperforming ODICE and QGPO overall. This indicates that the multimodal expressiveness of diffusion policies provides certain benefits for cross-town generalization. However, improving generative capability alone remains insufficient to fully address risk identification and action selection in new towns. ODICE achieves success rates below 10\% across all three traffic densities, indicating that traditional distribution correction methods may suffer from unstable density-ratio estimation when facing new state distributions. QGPO obtains low average rewards, even reaching -22.70 in the simple scenario, suggesting that purely generative policies without stable data-support constraints are vulnerable to distribution shift and accumulated closed-loop errors.

Overall, the Town05 generalization experiments show that the performance advantage of DiDrive is not limited to Town03. Compared with baseline methods that rely on conservative value constraints, behavior cloning constraints, or pure generative expressiveness, DiDrive maintains high success rates and high average rewards under new road environments and traffic distributions. This result supports the central claim of this paper: offline reinforcement learning for autonomous driving must not only address out-of-distribution value overestimation in the action space, but also improve the model's adaptability to key traffic interactions in new scenarios through risk-aware representation.

\subsection{Ablation Study}

\subsubsection{Training Analysis}

To understand the individual contributions of our proposed modules to training stability, we compare the learning curves of various ablated variants.

\begin{center}
\begin{minipage}{0.95\columnwidth}
    \centering
    \includegraphics[width=\linewidth]{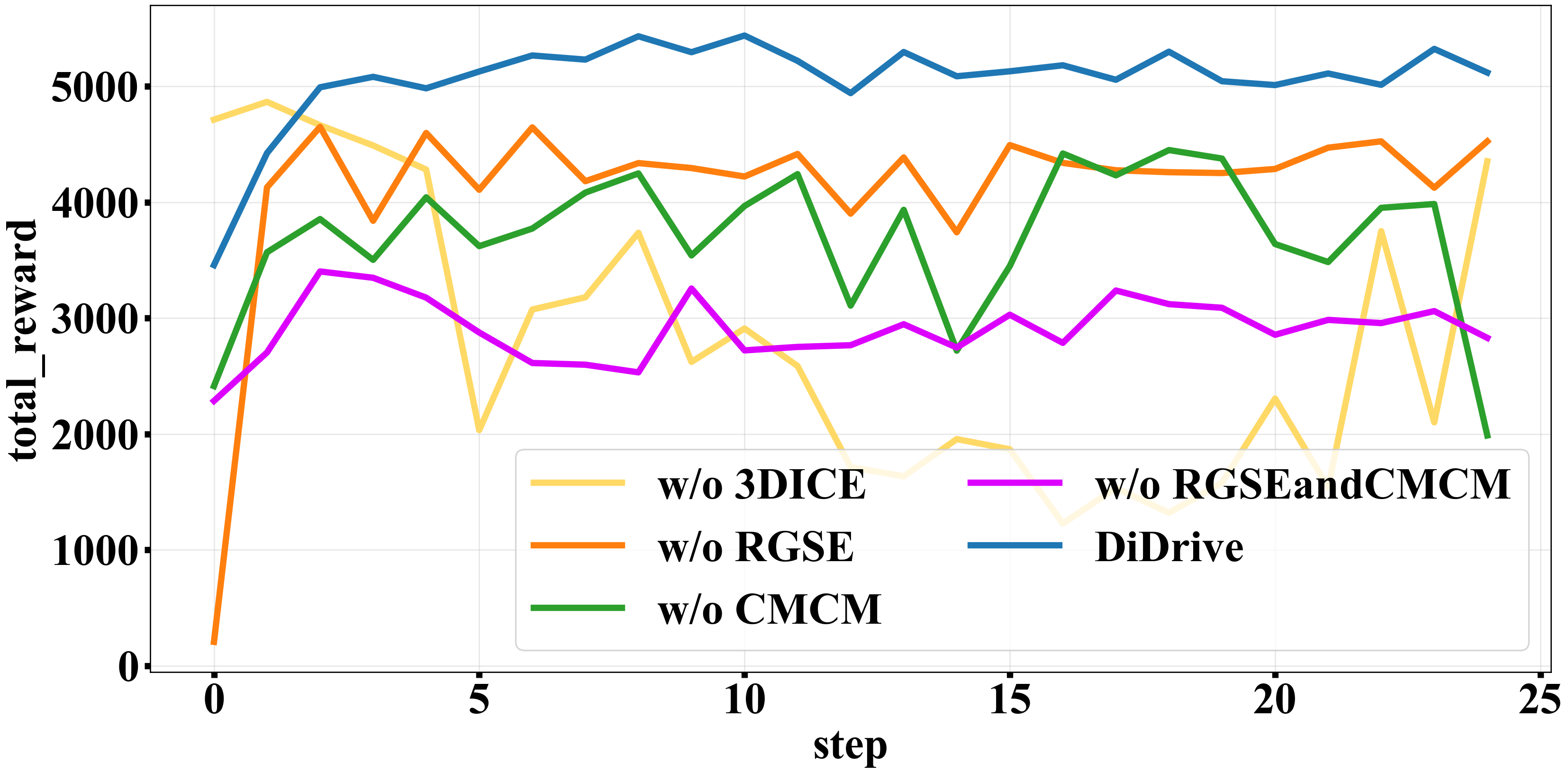}
    \captionof{figure}{Total reward analysis under different ablation settings.}
    \label{fig:ablation_reward}
\end{minipage}
\end{center}

\begin{center}
\begin{minipage}{0.95\columnwidth}
    \centering
    \includegraphics[width=\linewidth]{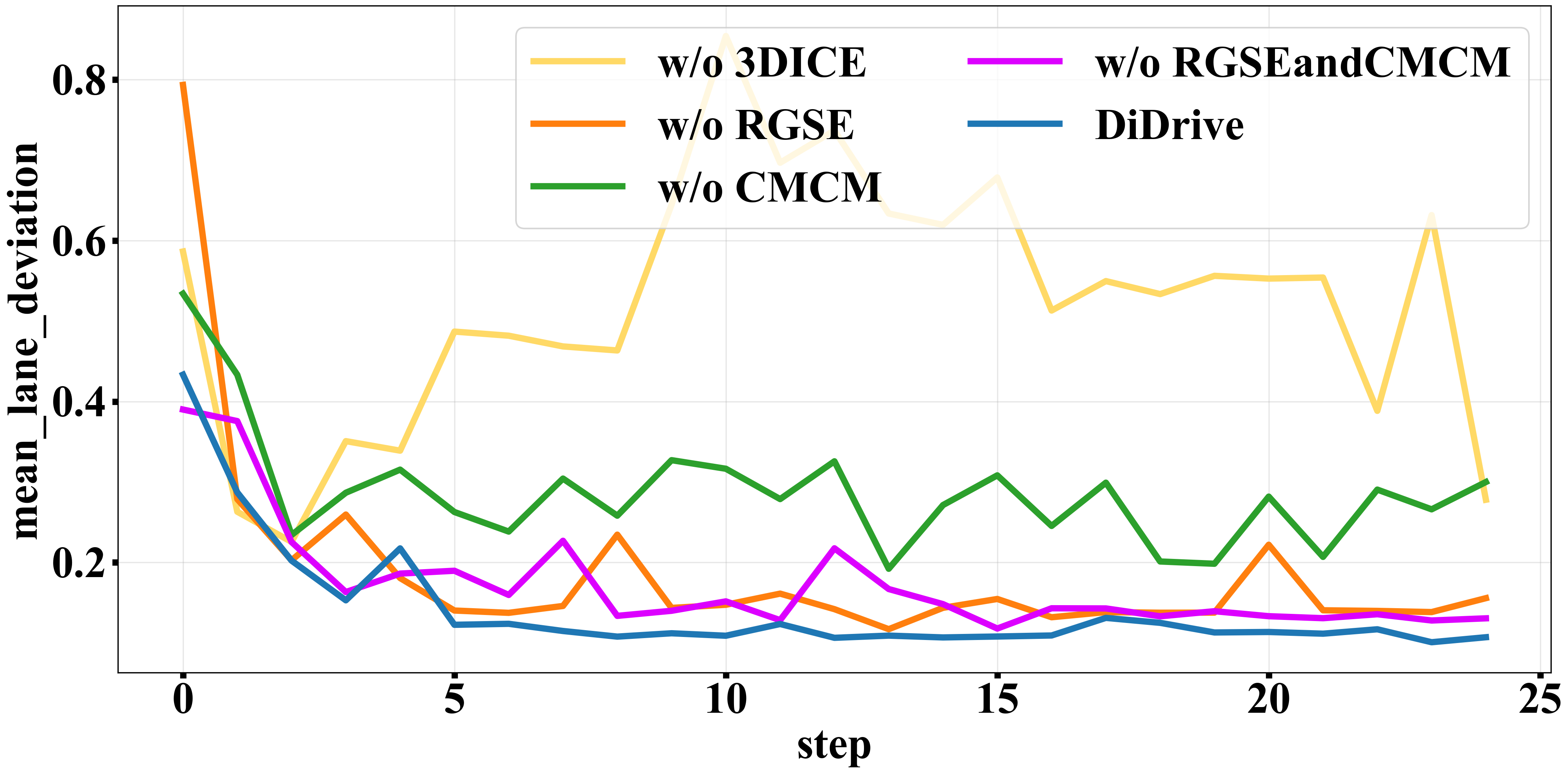}
    \captionof{figure}{Lane deviation analysis under different ablation settings.}
    \label{fig:ablation_deviation}
\end{minipage}
\end{center}

Figs.~\ref{fig:ablation_reward} and~\ref{fig:ablation_deviation} further reveal the synergistic mechanism among the components of DiDrive. The full model, represented by the blue curve, exhibits the most stable training trend: its reward rapidly converges to around 5200 in the early training stage, while the lane deviation remains stable at approximately 0.10 m. In sharp contrast, the variant without the 3DICE paradigm, represented by the yellow curve, suffers the most significant performance degradation. Its reward exhibits severe sawtooth-like oscillations, and the vehicle deviation frequently exceeds 0.60 m. This indicates that, without distribution correction guidance and stabilized optimization, the policy is easily affected by out-of-distribution action overestimation and heavy-tailed sample perturbations, making it difficult to maintain a stable driving trajectory.

The partial ablation of the representation modules further highlights the synergistic design of RHDif. When the CMCM module is removed, as depicted by the green curve, the policy exhibits pronounced lateral oscillations and severe reward fluctuations. This instability suggests that without high-level contextual filtering, the model's raw local risk responses are easily hijacked by background noise or pseudo-risks, leading to overreactive and erratic evasive maneuvers. Conversely, as shown by the orange curve, removing the RGSE module yields a relatively smooth lane-keeping trajectory, yet the total reward noticeably underperforms the full model. This divergence indicates a lack of sensitivity to critical local interactions; without RGSE's bottom-up feature enhancement, the agent struggles to swiftly identify and react to sudden, close-range threats like vehicle cut-ins, ultimately penalizing its overall driving efficiency and safety score. Finally, the basic variant lacking both RGSE and CMCM, represented by the purple curve, manages to maintain baseline lateral stability on straightforward road segments, but its total reward permanently stagnates at around 3000. This stark performance drop demonstrates that a standard diffusion backbone, stripped of hierarchical risk awareness, is fundamentally ill-equipped to handle fine-grained multi-vehicle interactions and complex, long-tailed traffic scenarios.

Ultimately, these results confirm that representation filtering and policy optimization are fundamentally interdependent. Only when RHDif's risk-calibrated perception is coupled with 3DICE's stable optimization can the model achieve a favorable balance between driving efficiency and safe control in complex traffic scenarios.

\subsubsection{Testing Analysis}

Beyond training dynamics, we further quantify how each decoupled component specifically contributes to the final closed-loop safety and driving efficiency.

\begin{table*}[t]
    \centering
    \caption{Ablation study of different components in the proposed framework.}
    \label{tab:ablation_study}
    \resizebox{\textwidth}{!}{
        \setlength{\tabcolsep}{4pt}
        \renewcommand{\arraystretch}{1.2}
        \begin{tabular}{c ccc ccc ccc ccc} 
            \toprule
            \multirow{2}{*}{\textbf{Comparison}} & \multicolumn{3}{c}{\textbf{Components}} & \multicolumn{3}{c}{\textbf{Simple Scenario (20 Vehicles)}} & \multicolumn{3}{c}{\textbf{Medium Scenario (40 Vehicles)}} & \multicolumn{3}{c}{\textbf{Complex Scenario (60 Vehicles)}} \\
            \cmidrule(lr){2-4} \cmidrule(lr){5-7} \cmidrule(lr){8-10} \cmidrule(lr){11-13}
             & 3DICE & RGSE & CMCM & Avg. Dist (m) $\uparrow$ & Avg. Reward $\uparrow$ & SR $\uparrow$ & Avg. Dist (m) $\uparrow$ & Avg. Reward $\uparrow$ & SR $\uparrow$ & Avg. Dist (m) $\uparrow$ & Avg. Reward $\uparrow$ & SR $\uparrow$ \\
            \midrule
            A & & \checkmark & \checkmark & 254 & 3913.01 & 73\% & 239 & 3645.50 & 66\% & 213 & 3313.04 & 44\% \\
            B & \checkmark & & & 284 & 3736.12 & 86\% & 260 & 3629.97 & 83\% & 241 & 3385.79 & 78\% \\
            C & \checkmark & \checkmark & & 262 & 3866.46 & 77\% & 254 & 3852.53 & 76\% & 235 & 3594.67 & 67\% \\
            D & \checkmark & & \checkmark & \textbf{288} & 4407.93 & 90\% & 272 & 4211.11 & 84\% & 245 & 3754.90 & 80\% \\
            E & \checkmark & \checkmark & \checkmark & 286 & \textbf{4902.94} & \textbf{92\%} & \textbf{299} & \textbf{4257.76} & \textbf{86\%} & \textbf{254} & \textbf{4295.68} & \textbf{85\%} \\
            \bottomrule
        \end{tabular}
    }
\end{table*}

By constructing five variants, we further evaluate the individual contributions of 3DICE, RGSE, and CMCM in DiDrive. The quantitative results under three traffic densities are summarized in Table~\ref{tab:ablation_study}. Overall, the performance improvement of DiDrive does not come from a single component, but relies on the coordination between the policy optimization mechanism and the risk-aware representation module.

First, the 3DICE paradigm, as indicated by the comparison between Row A and Row E, serves as the core foundation for stable system optimization. After removing this module, the performance drops significantly across all densities: for instance, the success rate falls from 86\% to 66\% in the medium scenario, and from 85\% to 44\% in the complex scenario, where the average reward also decreases from 4295.68 to 3313.04. This indicates that relying solely on the diffusion backbone is insufficient to handle long-tail risks and distribution shifts in high-dimensional driving decisions. Without in-sample guidance and support-set constraints, the generated actions are more susceptible to out-of-distribution value overestimation, making the policy unstable in complex interactive environments.

Second, RGSE and CMCM play different roles in the spatiotemporally decoupled hierarchical diffusion architecture: RGSE focuses on amplifying local risk responses, while CMCM provides global semantic filtering to calibrate these signals. With 3DICE retained, introducing only RGSE, as shown in Row C, yields success rates of 76\% and 67\% in the medium and complex scenarios, with average driving distances of 254 m and 235 m, respectively. These metrics fall noticeably short of the 83\% and 78\% success rates, alongside the 260 m and 241 m average driving distances achieved by the baseline without any representation module in Row B. This result demonstrates that enhancing local risk responses alone does not necessarily lead to better performance. Without the global semantic filtering provided by CMCM, RGSE may overly amplify the local responses of non-critical vehicles or background objects, leading to behaviors such as excessive avoidance, frequent sudden braking, or conservative stagnation.

In contrast, the variant with only CMCM added, as shown in Row D, performs more stably across varying densities. It achieves an average reward of 4211.11 in the medium scenario and a success rate of 80\% with a 3754.90 reward in the complex scenario, generally outperforming both Row B and Row C. This suggests that cross-modal contextual modulation can suppress background interference and filter local pseudo-risk responses. However, without the low-level risk enhancement provided by RGSE, the model's sensitivity to sudden dangerous interactions remains limited, and its performance is therefore still lower than that of the full model.

After simultaneously introducing RGSE and CMCM, the full model in Row E achieves the highest performance across all metrics. Notably, it reaches an 86\% success rate and a 299 m driving distance in the medium scenario, while improving the success rate in the complex scenario to 85\% and achieving the highest average reward of 4295.68. This result indicates that RGSE provides low-level enhancement for local risk interactions, while CMCM performs global semantic correction on local high-response features. By combining the two, the model can focus on key objects that may compress the safety boundary while reducing interference from background redundancy and pseudo-risk signals, thereby maintaining more stable decision-making in complex vehicle interactions and dense traffic environments.

\subsubsection{Risk-Aware Module Analysis}

To further examine the functional roles of the proposed risk-aware modules, we introduce additional diagnostic metrics beyond standard closed-loop evaluation. While success rate and collision rate reflect final driving outcomes, they do not directly reveal how each module contributes to risk perception and feature organization. Therefore, we first analyze the internal behavior of the CMCM module, focusing on whether the local risk signals amplified by RGSE can be effectively organized, filtered, and aligned with the diagnostic reference risk distribution under global contextual semantics. To this end, we adopt four evaluation metrics: Risk Gate Activation and Risk Attention Alignment as internal diagnostic metrics, alongside Collision Rate and Success Rate as closed-loop indicators.

\textbf{Risk Gate Activation:} This metric measures the average intensity of the risk-gated feature response, defined as:
\begin{equation}
\text{Risk Gate Activation} = \mathbb{E}[\text{risk\_gate} - 0.5],
\end{equation}
where $\text{risk\_gate}=\sigma(\text{MLP}(\mathbf{r}_{risk}))+0.5$. Here, $\mathbf{r}_{risk}$ denotes the risk-related input vector, including ego speed, lane deviation, front-vehicle distance, and relative velocity. A higher value indicates stronger activation of risk-related modalities or channels. Since both the full model and the w/o CMCM variant contain RGSE, this metric reveals whether the risk features amplified by RGSE can be effectively preserved after contextual modulation.

\textbf{Risk Attention Alignment:} This metric further measures whether the learned risk-aware attention is consistent with a reference risk distribution derived from driving-state indicators. It is computed as the cosine similarity between the learned modal risk weights and the reference risk distribution:
\begin{equation}
\text{Risk Attention Alignment}
=
\cos(\mathbf{w}_{learned},\mathbf{r}_{true}).
\end{equation}
Here, $\mathbf{w}_{learned}$ is obtained from the combination of the RGSE risk activation and the CMCM spatial gate. The reference vector $\mathbf{r}_{true}$ is constructed only for diagnostic evaluation and is not used for model training or action selection.

Specifically, $\mathbf{r}_{true}$ is computed from normalized rule-based risk indicators associated with different input modalities:
\begin{equation}
\mathbf{r}_{true}
=
\text{Normalize}
\left[
r_{ego},
r_{lane},
r_{lidar},
r_{veh},
r_{wp}
\right],
\end{equation}
where $r_{ego}$ denotes the ego-state risk derived from excessive speed or acceleration, $r_{lane}$ denotes the lane-deviation risk derived from the absolute lateral offset, $r_{lidar}$ denotes the obstacle-proximity risk derived from the minimum LiDAR distance, $r_{veh}$ denotes the surrounding-vehicle interaction risk derived from front-vehicle distance and relative velocity, and $r_{wp}$ denotes the waypoint-tracking risk derived from the deviation between the current motion direction and the navigation waypoint direction. The normalization operation rescales these indicators into a comparable distribution across modalities:
\begin{equation}
\text{Normalize}(\mathbf{x})=\frac{\mathbf{x}}{\sum_{m}x_m+\epsilon},
\end{equation}
where $\epsilon$ is a small constant for numerical stability. A higher alignment score indicates that the learned attention focuses more consistently on the modalities associated with higher reference risk rather than irrelevant background responses. Therefore, this metric is used only to analyze the internal behavior of RGSE and CMCM, rather than serving as an optimization objective.

\begin{center}
\begin{minipage}{0.95\columnwidth}
    \centering

    \begin{minipage}[b]{0.48\linewidth}
        \centering
        \includegraphics[width=\linewidth]{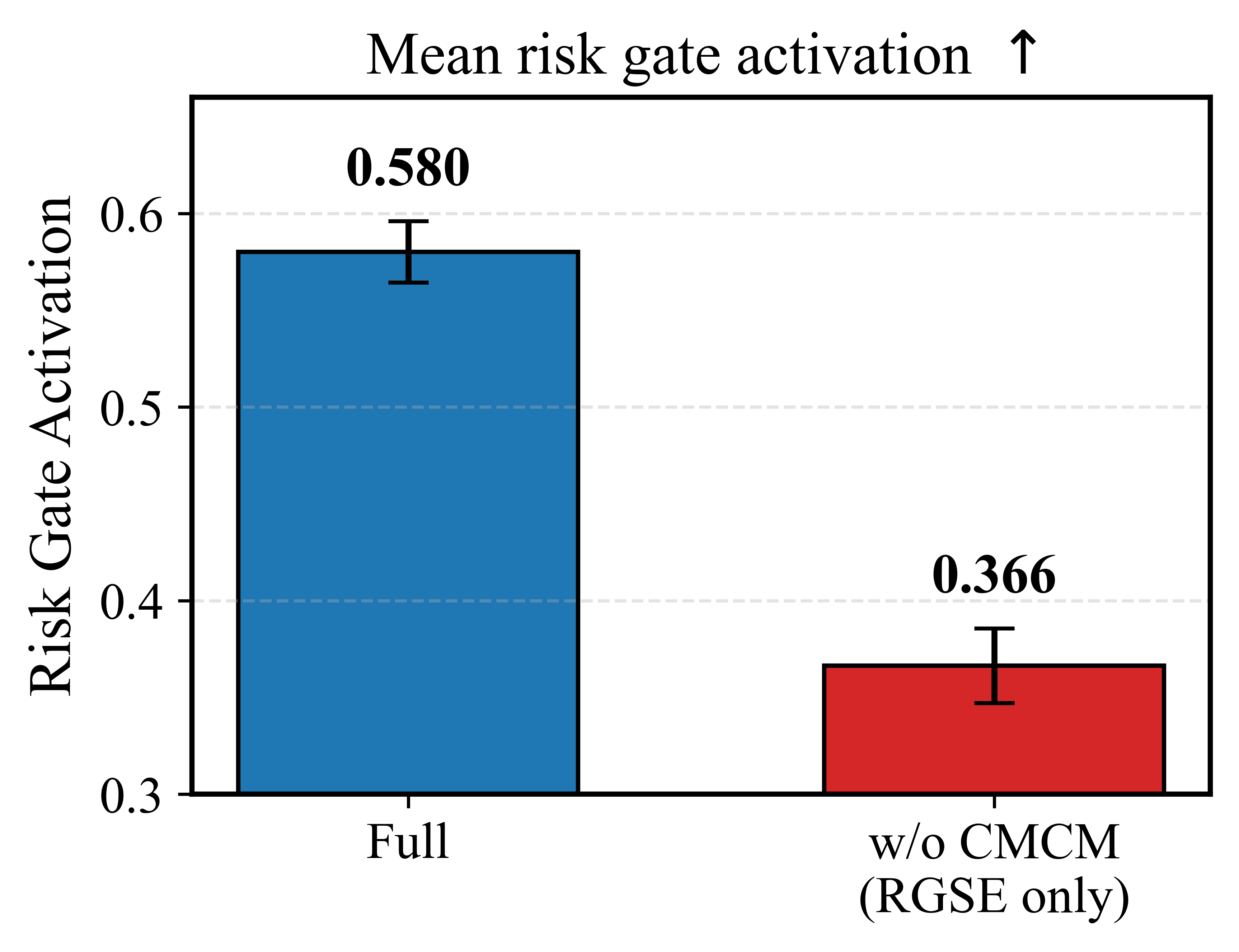}
        
        \small (a) Risk Gate Activation
    \end{minipage}
    \hfill
    \begin{minipage}[b]{0.48\linewidth}
        \centering
        \includegraphics[width=\linewidth]{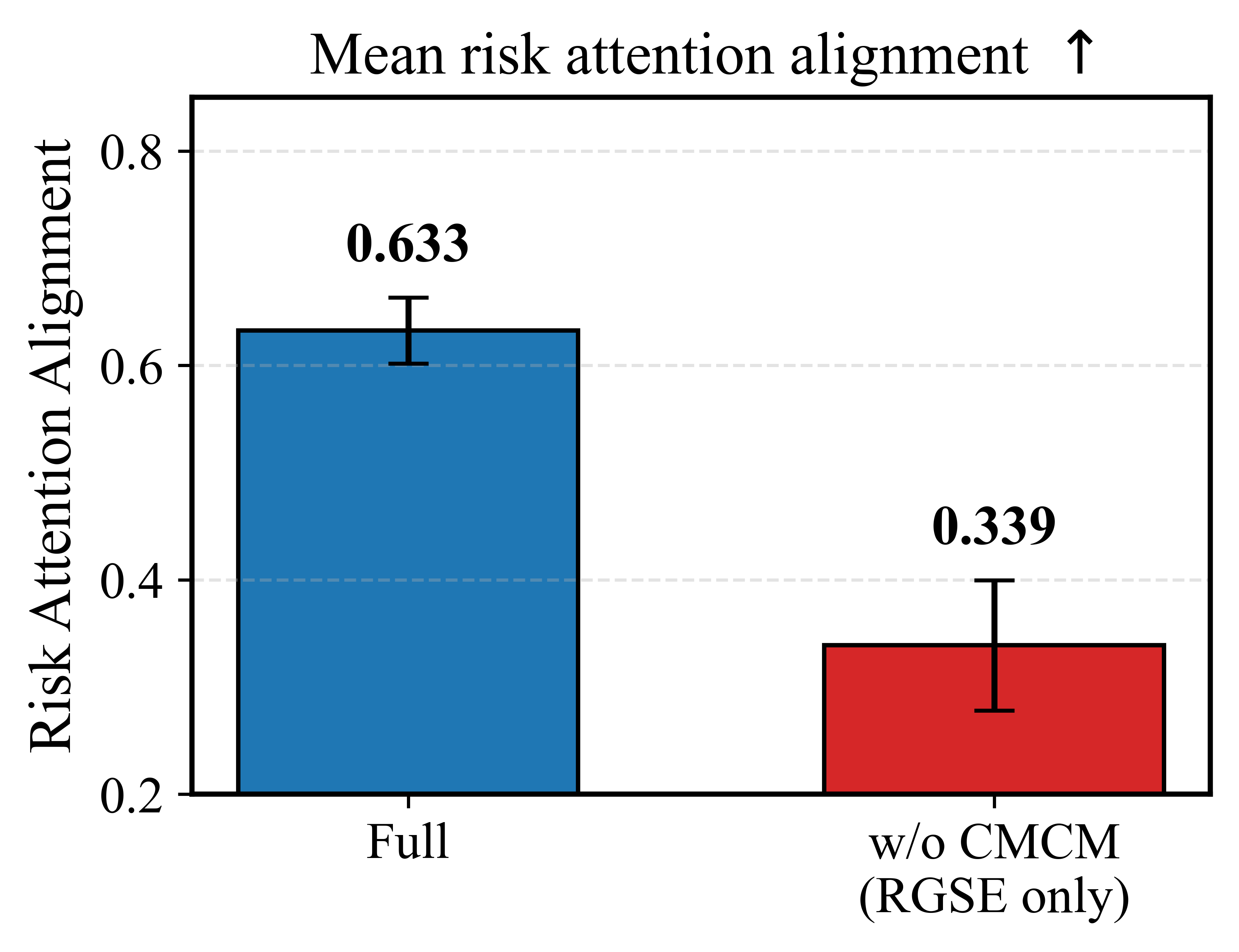}
        
        \small (b) Risk Attention Alignment
    \end{minipage}

    \vspace{0.6em}

    \begin{minipage}[b]{0.48\linewidth}
        \centering
        \includegraphics[width=\linewidth]{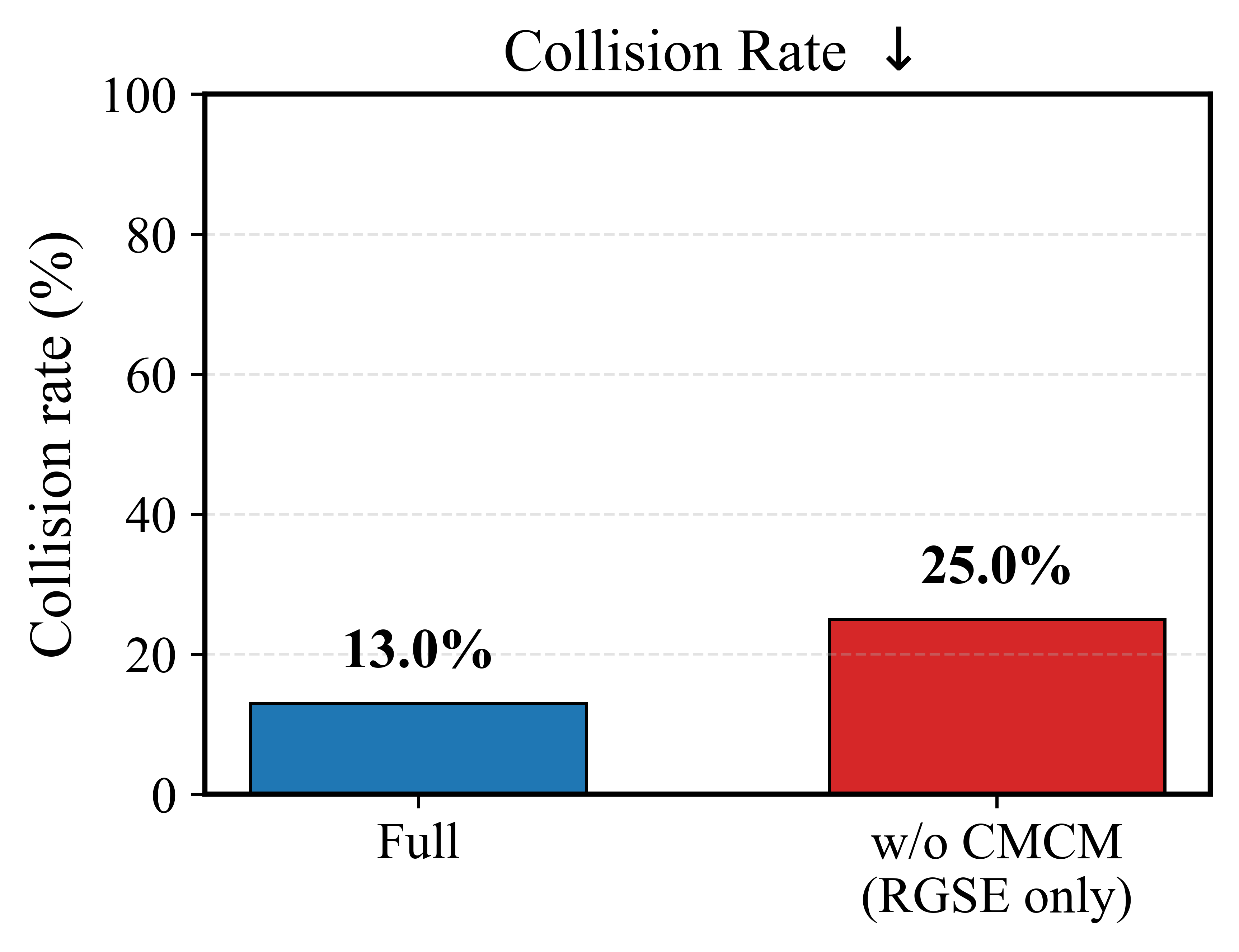}
            
            \small (c) Collision Rate
        \end{minipage}
        \hfill
        \begin{minipage}[b]{0.48\linewidth}
            \centering
            \includegraphics[width=\linewidth]{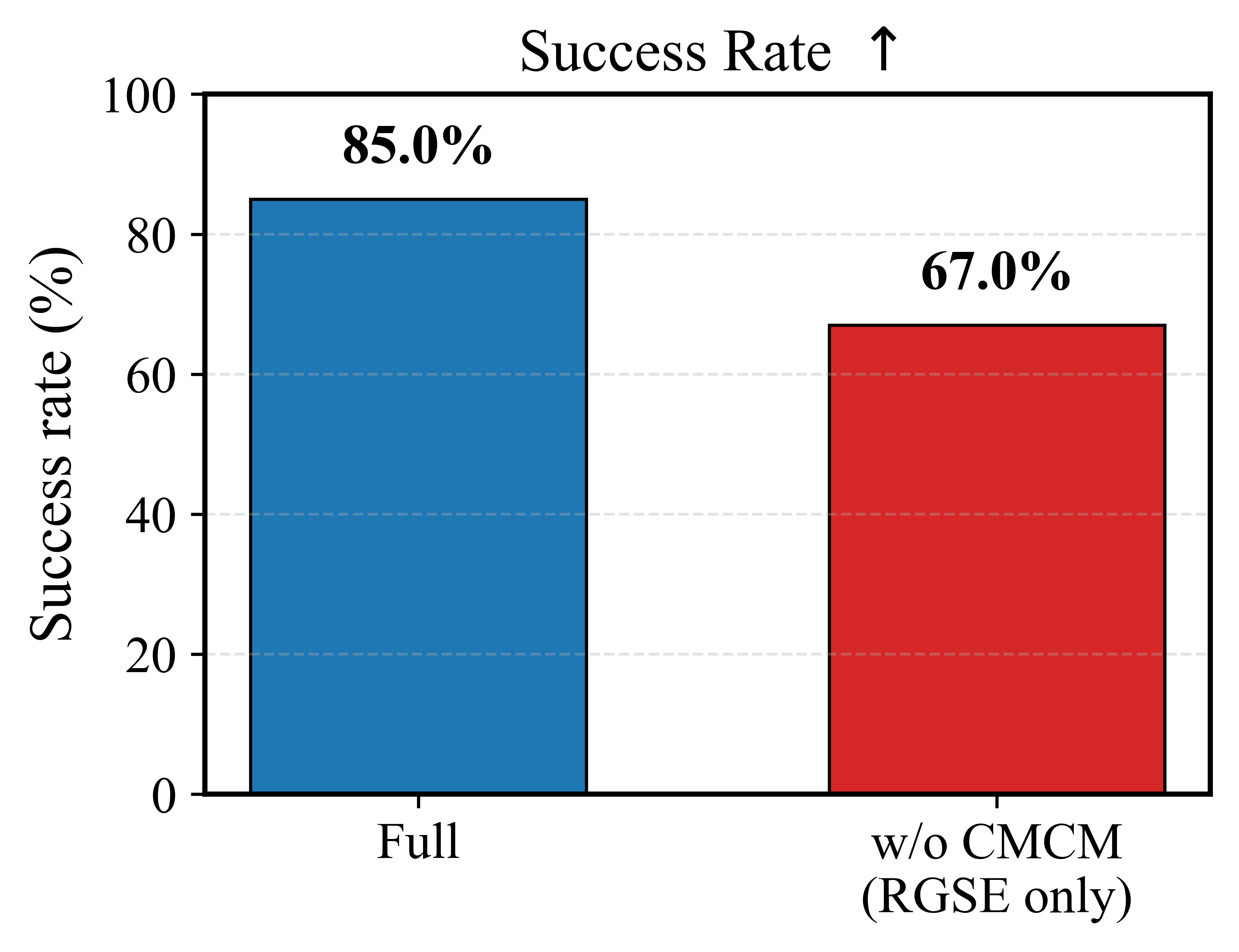}
            
            \small (d) Success Rate
        \end{minipage}
    
        \captionof{figure}{Effectiveness analysis of CMCM in complex scenarios.}
        \label{fig:cmcm_analysis}
    \end{minipage}
    \end{center}
    
As shown in Fig.~\ref{fig:cmcm_analysis}, the comparison between the full model and the w/o CMCM variant reveals the role of CMCM in risk-aware feature calibration. Since RGSE is retained in both models, the difference lies in whether the amplified local risk features are further processed by cross-modal contextual modulation. Risk Gate Activation reflects the strength of risk-aware responses, but higher activation alone does not necessarily imply better safety, as misaligned activation may correspond to false risk responses. Therefore, it must be interpreted alongside Risk Attention Alignment.

The improvement in Risk Attention Alignment indicates that CMCM successfully aligns learned attention with the diagnostic reference risk distribution. Rather than indiscriminately amplifying all risk-related features, CMCM selectively filters background interference to preserve semantically meaningful risk cues. This top-down contextual calibration enables the policy to distinguish genuine threats---such as close leading vehicles or high-risk surrounding agents---from irrelevant background noise. Consequently, the closed-loop metrics further validate this role: the integration of CMCM yields a higher success rate and a lower collision rate, confirming that improved risk-attention alignment translates into safer driving behavior.

Next, we examine the RGSE module to verify whether the model can perceive local safety-critical interactions and respond properly under high-risk conditions. While CMCM focuses on global calibration, RGSE is designed to enhance bottom-up local risk perception. Thus, the RGSE ablation focuses on whether removing this module weakens the policy's capacity to identify near-collision states and suppress unsafe acceleration. We adopt four evaluation metrics: TTC $<2.5$ s Rate and High-risk Throttle Mean as internal diagnostic metrics, alongside Collision Rate and Success Rate as closed-loop indicators.

\textbf{TTC $<2.5$ s Rate:} The proportion of evaluation timesteps where the Time-To-Collision (TTC) falls below $2.5$ seconds. TTC estimates the time remaining before a potential rear-end collision at the current relative velocity. We adopt the 2.5-second threshold because it aligns with standard human perception-reaction times in traffic engineering. Falling below this margin effectively marks a high-risk, near-collision state; therefore, a lower rate indicates safer temporal margins.

\textbf{High-risk Throttle Mean:} The average throttle output during high-risk states. In our evaluation, a high-risk state is triggered either when the temporal safety margin is critically low ($\text{TTC} < 2.5 \text{ s}$) or when a composite environmental risk score (integrating factors like front-vehicle distance, lane deviation, and LiDAR obstacles) exceeds $0.5$. A lower mean throttle indicates that the policy has learned to effectively suppress acceleration when potential danger is detected, reflecting a highly risk-sensitive control strategy integrated into the training paradigm.

\begin{center}
\begin{minipage}{0.95\columnwidth}
    \centering

    \begin{minipage}[b]{0.48\linewidth}
        \centering
        \includegraphics[width=\linewidth]{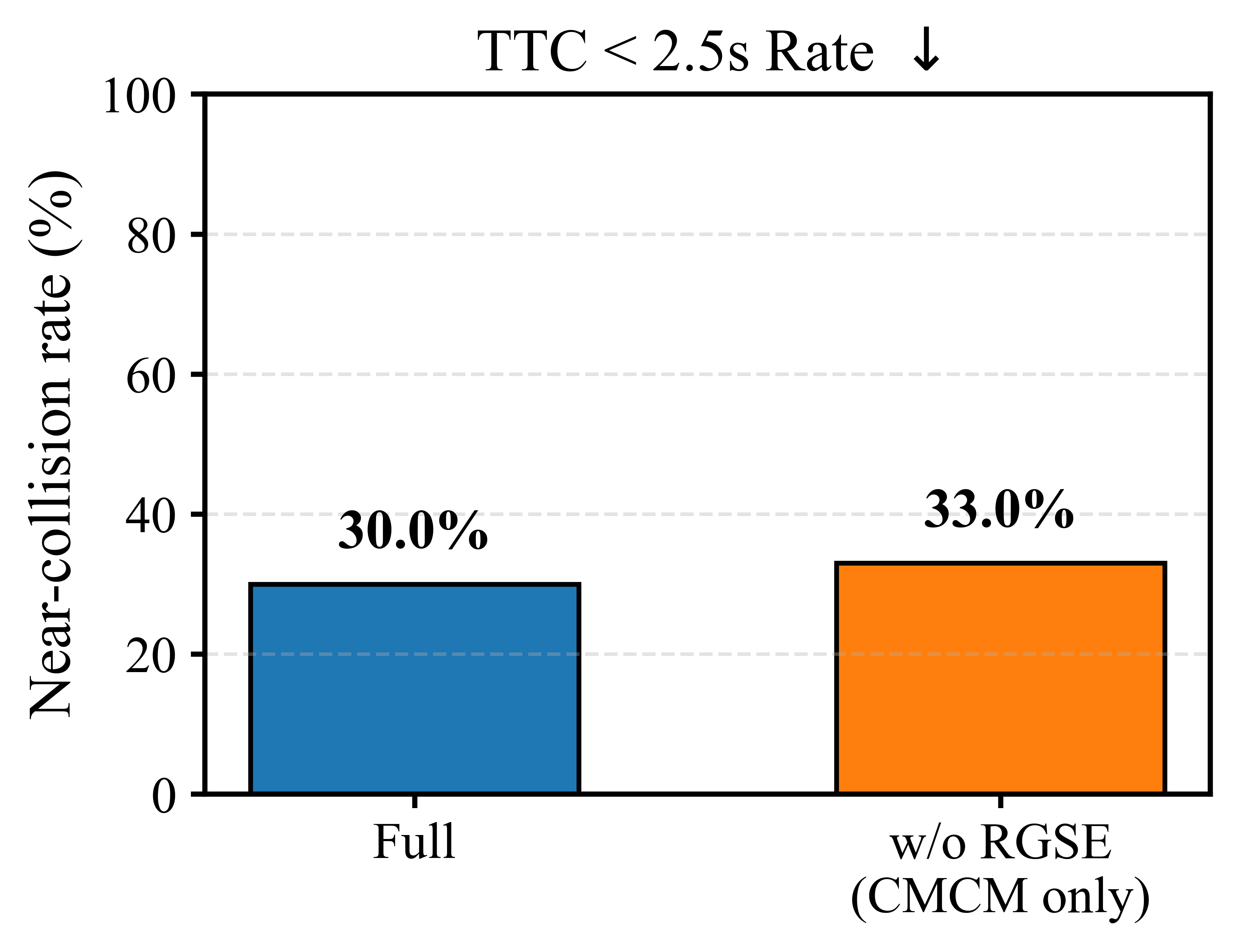}
        
        \small (a) TTC $<2.5$ s Rate
    \end{minipage}
    \hfill
    \begin{minipage}[b]{0.48\linewidth}
        \centering
        \includegraphics[width=\linewidth]{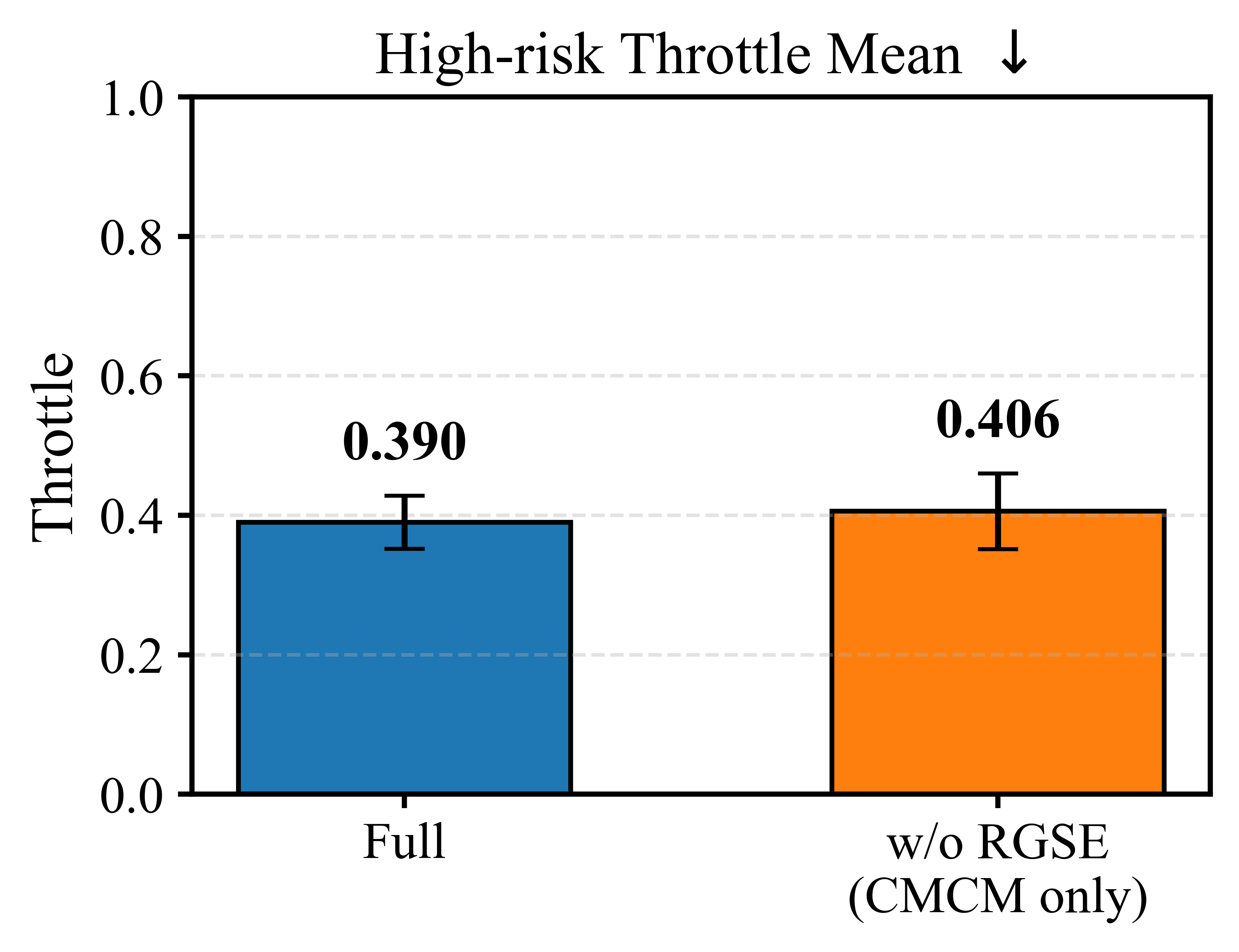}
        
        \small (b) High-risk Throttle Mean
    \end{minipage}

    \vspace{0.6em}

    \begin{minipage}[b]{0.48\linewidth}
        \centering
        \includegraphics[width=\linewidth]{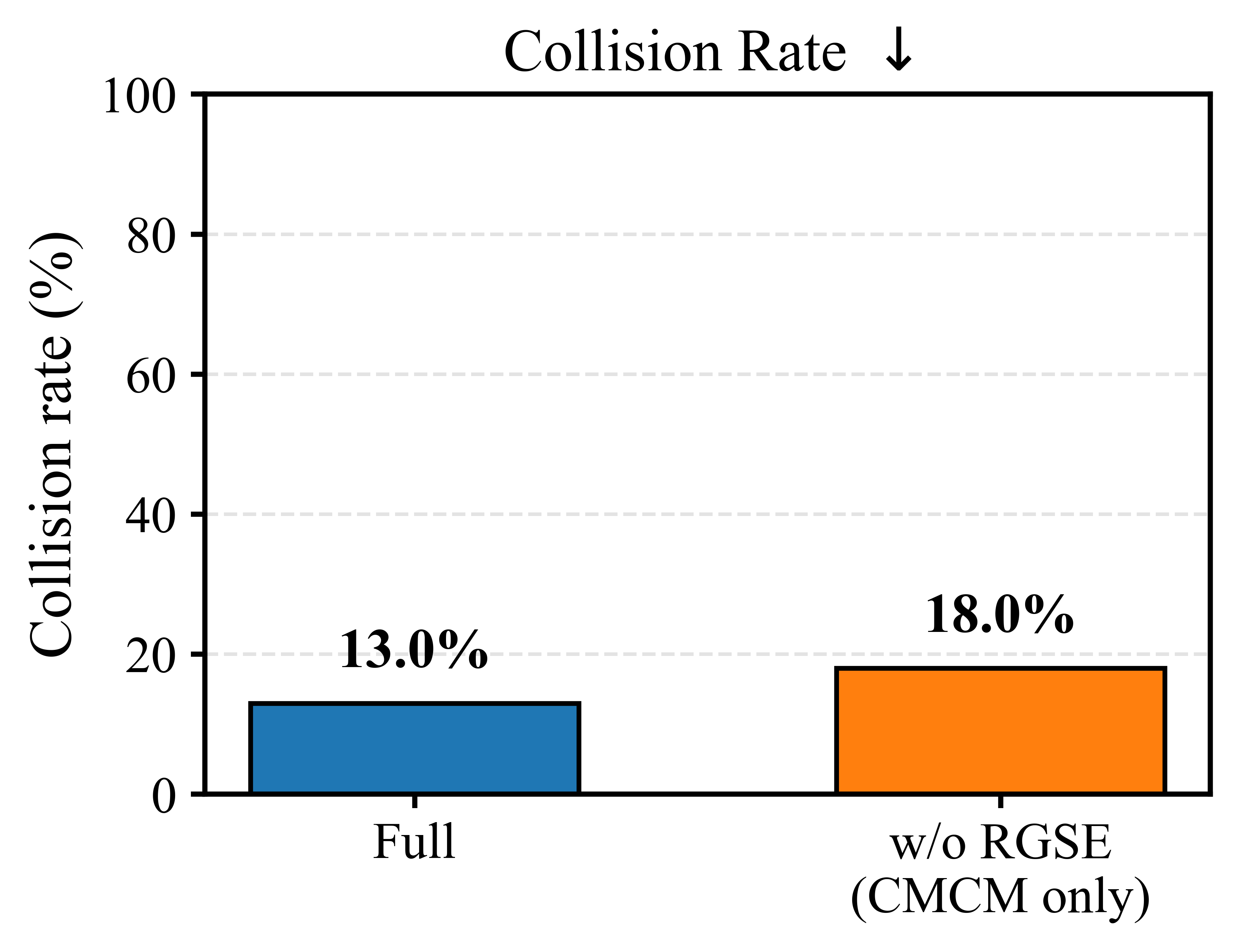}
        
        \small (c) Collision Rate
    \end{minipage}
    \hfill
    \begin{minipage}[b]{0.48\linewidth}
        \centering
        \includegraphics[width=\linewidth]{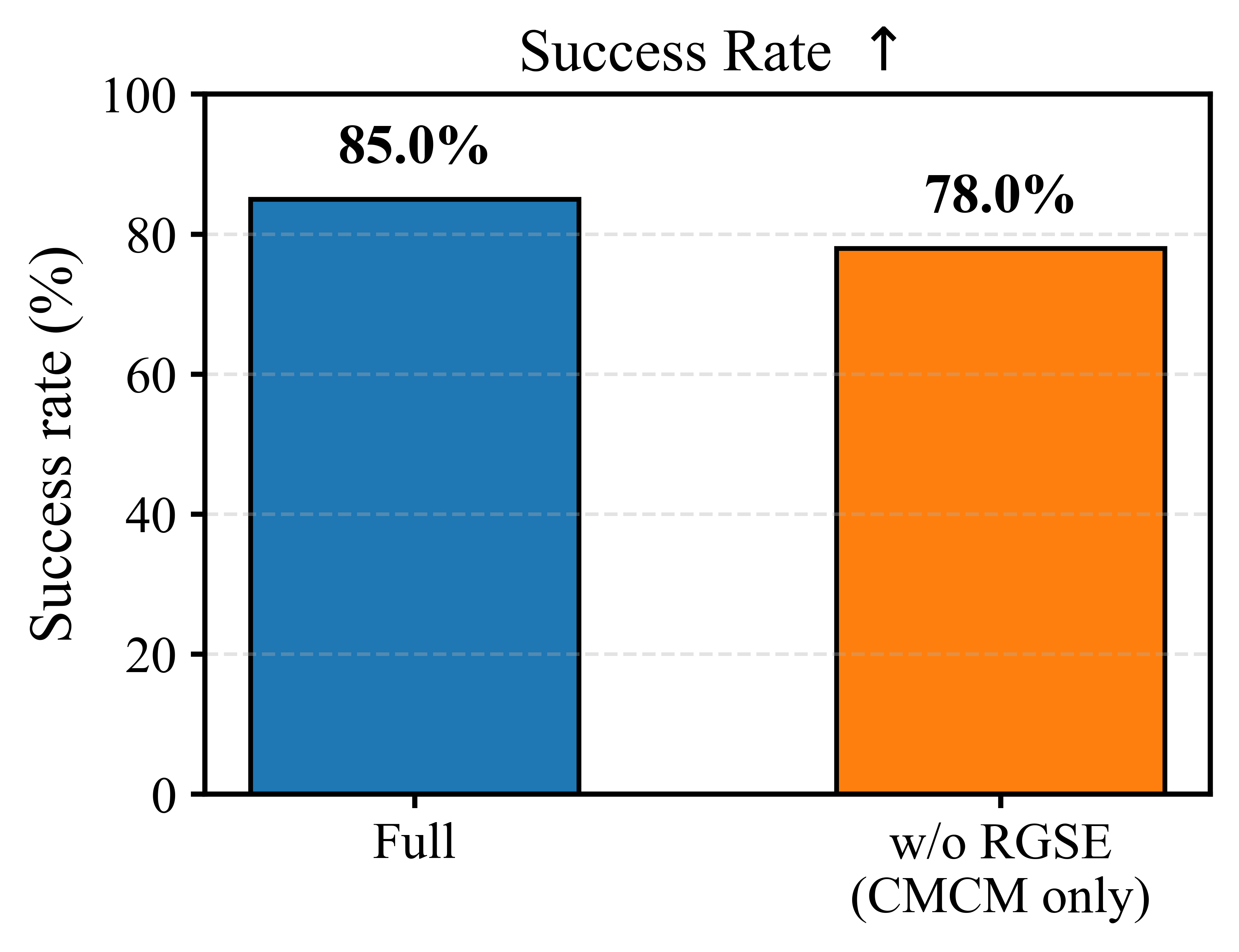}
        
        \small (d) Success Rate
    \end{minipage}

    \captionof{figure}{Effectiveness analysis of RGSE in complex scenarios.}
    \label{fig:rgse_analysis}
\end{minipage}
\end{center}

As shown in Fig.~\ref{fig:rgse_analysis}, removing RGSE leads to degradation in both diagnostic risk-aware metrics and closed-loop safety performance. Compared with the full model, the w/o RGSE variant shows a higher TTC $<2.5$ s Rate, increasing from 30.0\% to 33.0\%. This indicates that the ego vehicle enters near-collision states more frequently when RGSE is removed. Since TTC directly reflects the temporal margin before a potential rear-end collision, this result suggests that CMCM alone is insufficient to capture local interaction risks such as close leading vehicles, sudden cut-ins, and short-distance traffic conflicts.

The High-risk Throttle Mean further reveals the control-level effect of removing RGSE. The w/o RGSE variant produces a higher throttle output under high-risk states, increasing from 0.390 to 0.406. Although the numerical difference is moderate, it indicates that the policy without RGSE is less capable of suppressing acceleration when potential hazards are present. In dense traffic scenarios, even a small increase in throttle under risky conditions may reduce the available reaction time and increase the probability of entering near-collision states. This result shows that RGSE contributes not only to risk feature extraction but also to safer control behavior by encouraging the policy to reduce acceleration under dangerous conditions.

The closed-loop metrics provide further evidence for the effectiveness of RGSE. Without RGSE, the collision rate increases from 13.0\% to 18.0\%, while the success rate decreases from 85.0\% to 78.0\%. These results are consistent with the diagnostic metrics: weaker local risk perception leads to more frequent near-collision exposure and less effective throttle suppression, which eventually results in more collisions and fewer successful episodes. Therefore, RGSE plays a crucial role in enhancing the policy's sensitivity to local safety-critical interactions.

Overall, these results demonstrate that RGSE is responsible for bottom-up local risk amplification in DiDrive. By dynamically enhancing spatiotemporal features related to surrounding vehicles, front-distance risks, lane deviations, and obstacle proximity, RGSE enables the policy to identify hazardous situations earlier and suppress unsafe acceleration more effectively. In contrast, removing RGSE weakens the model's ability to perceive local interaction risks, making the policy more likely to enter near-collision states and increasing the probability of collision. Together with the CMCM analysis, these results show that RGSE and CMCM play complementary roles: RGSE enhances sensitivity to local safety-critical cues, while CMCM calibrates these risk responses through global contextual semantics. Their combination enables DiDrive to achieve safer and more robust decision-making in complex traffic scenarios.

\section{Conclusion}

This paper addresses training instability in offline reinforcement learning for autonomous driving caused by distribution shift, high-dimensional traffic redundancy, and heavy-tailed risk signals. To tackle these challenges, we proposed DiDrive, a risk-aware hierarchical diffusion framework. By synergizing the 3DICE policy optimization paradigm with the RHDif representational architecture, DiDrive performs risk-aware policy redistribution within the support of real driving data, substantially decreasing the reliance on extrapolated value estimation for out-of-distribution actions.

Extensive closed-loop evaluations in the high-fidelity CARLA simulator demonstrate the superiority of our approach. In high-density complex traffic scenarios involving 60 background vehicles, DiDrive achieved the highest success rate of 85\% and a peak average reward of 4295.68, significantly outperforming state-of-the-art baselines such as Diffusion-QL and CQL, which achieved success rates of 77\% and 74\% respectively. Furthermore, in zero-shot cross-town generalization tests, DiDrive maintained a robust 81\% success rate in complex configurations. These quantitative results prove that our risk-aware hierarchical representation successfully extracts generalized safety-critical features rather than overfitting to specific map structures.

Despite these advancements, this work has certain limitations. First, while the framework stabilizes training, the multi-step reverse denoising of diffusion models inherently introduces inference latency, posing challenges for high-frequency real-time deployment on vehicle-embedded platforms. Second, our current evaluations are conducted within a simulator, leaving the simulation-to-reality domain gap unexplored. Future work will focus on integrating accelerated sampling techniques, such as consistency models, to reduce inference latency. Additionally, we plan to validate the framework on real-world autonomous driving datasets and explore continuous simulation-to-reality transfer strategies for physical vehicle deployment.

\printcredits

\bibliographystyle{unsrt}

\bibliography{main}



\end{document}